\documentclass{article}

\usepackage{microtype}
\usepackage{graphicx}
\usepackage{subcaption}
\usepackage{booktabs} 

\usepackage{hyperref}

\usepackage[accepted]{icml2026}

\usepackage{amsmath}
\usepackage{amssymb}
\usepackage{mathtools}
\usepackage{amsthm}
\usepackage{multirow}

\DeclareMathOperator*{\argmax}{arg\,max}

\usepackage[capitalize,noabbrev]{cleveref}

\theoremstyle{plain}

\theoremstyle{definition}

\theoremstyle{remark}

\usepackage[textsize=tiny]{todonotes}

\usepackage[english]{babel}
\usepackage[autostyle, english = american]{csquotes}
\MakeOuterQuote{"}

\icmltitlerunning{Target-Aware Bandit Allocation for Scalable Surrogate Optimization in Chemical Space}

\begin{document}

\twocolumn[
  \icmltitle{Target-Aware Bandit Allocation for \\ Scalable Surrogate Optimization in Chemical Space}



  \icmlsetsymbol{equal}{*}

  \begin{icmlauthorlist}
      \icmlauthor{Mohammad Haddadnia}{dfci,hms,chan}
      \icmlauthor{Yuvan Chali}{dfci,hms,yale}
      \icmlauthor{Abhilash Jayaraj}{dfci,hms}
      \icmlauthor{Constance Kraay}{hms-grad}
      \icmlauthor{Joana Reis}{dfci,hms}

      \icmlauthor{Felix Strieth-Kalthoff}{buw,buwizmd}
      \icmlauthor{Haribabu Arthanari}{dfci,hms}
  \end{icmlauthorlist}

  \icmlaffiliation{dfci}{Dana-Farber Cancer Institute, Boston, MA, USA}
  \icmlaffiliation{hms}{Department of Biological Chemistry and Molecular Pharmacology, Harvard Medical School, Boston, MA, USA}
  \icmlaffiliation{chan}{Department of Biostatistics, Harvard T.H. Chan School of Public Health, Boston, MA, USA}
  \icmlaffiliation{yale}{Department of Chemistry, Yale University, New Haven, CT, USA}
  \icmlaffiliation{hms-grad}{Graduate Program in Biophysics, Harvard University, Cambridge, MA, USA}
  \icmlaffiliation{buw}{University of Wuppertal, School of Mathematics and Natural Sciences, Wuppertal, Germany}
  \icmlaffiliation{buwizmd}{University of Wuppertal, Interdisciplinary Center of Machine Learning and Data Analytics, Wuppertal, Germany}

  \icmlcorrespondingauthor{Felix Strieth-Kalthoff}{strieth-kalthoff@uni-wuppertal.de}
  \icmlcorrespondingauthor{Haribabu Arthanari}{hari\_arthanari@hms.harvard.edu}

  \icmlkeywords{Machine Learning, ICML}

  \vskip 0.3in
]



\printAffiliationsAndNotice{}  

\begin{abstract}
Identifying high-utility candidates from massive discrete spaces under expensive evaluations is a recurring challenge across the sciences, with structure-based drug discovery as a prominent example.
While surrogate-based optimization can increase sample efficiency by reducing the number of expensive evaluations, modern molecular libraries have reached billions to trillions of compounds, making full-library surrogate inference itself a major computational bottleneck. 
We introduce \textsc{BoBa}, a bandit-guided surrogate optimization framework that eliminates full-library inference by adaptively allocating computation across partitions of the action space. 
By treating partitions as arms in a multi-armed bandit, \textsc{BoBa} concentrates inference and evaluations on empirically promising partitions while maintaining principled exploration. 
Experiments on real-world synthesis-on-demand libraries demonstrate that optimism-under-uncertainty bandits, combined with meaningful action space partitioning, are essential for effective allocation of inference and evaluations. 
Our findings reveal a tunable tradeoff between screening performance and surrogate inference cost, which supports practical optimization over current libraries, and establishes a viable route to ultra-large library virtual screening. 
\end{abstract}

\section{Introduction}

The earliest stages of drug discovery are dominated by a large and resource-intensive search problem: finding molecules that potently and selectively engage a biological target, and can be advanced into therapeutic leads. 
This hit identification stage remains a major bottleneck in the drug discovery pipeline, both in terms of cost and time. 
Computational approaches offer a fast and inexpensive route to exploring diverse sets of candidates, and recent advances in algorithms, computing, and ultra-large compound libraries have renewed optimism about their impact \citep{Lyu2019, Gorgulla2020, Grygorenko2020, Sadybekov2021, Eisenhuth2025}. 
Importantly, costs increase sharply at each downstream stage of the drug discovery pipeline \citep{Morgan2011}, which makes the quality of early computational decisions disproportionately critical.

Computational candidate selection in drug discovery is fundamentally a discrete, sequential decision-making problem. 
Although the space of possible molecules is astronomically large, with estimates ranging from $10^{60}$–$10^{200}$ \citep{Restrepo2022}, only a small fraction can be practically synthesized from commercially available building blocks at reasonable cost \citep{Papidocha2026}. 
In silico drug discovery therefore operates over large but finite libraries of synthetically accessible compounds \citep{Shoichet2004}, framing hit identification as a search over a large but discrete action space under budget constraints. 

Active Learning \citep{Settles2012} and Bayesian Optimization \citep{Garnett2012, Garnett2015, Jiang2018, garnett_bayesoptbook_2023} have become central tools for navigating these settings. 
While libraries have historically been evaluated exhaustively using molecular docking, a physics-informed simulation that estimates binding affinity between a molecule and a biological target, this strategy became prohibitively expensive as library sizes grew to billions of molecules.
At the same time, recent studies have underscored the importance of screening larger and more structurally diverse libraries to increase true-hit rates and potency \citep{Liu2025, Lyu2023, Gloriam2019}.
This need motivated surrogate-based optimization techniques that evaluate only a small subset of candidates and use learned models to guide subsequent selection \citep{Reker2015, PyzerKnapp2018, Reker2019, Graff2021}.

Recent years have witnessed yet another regime shift. 
Advances in make-on-demand chemistry have expanded accessible libraries from billions to trillions of molecules \citep{enamine, Hoffmann2019, Warr2022, Gorgulla2023}, altering the computational cost structure.
At these scales, cost is no longer dominated solely by expensive physics-informed evaluations. 
Instead, inference over the candidate set itself becomes a bottleneck: even a single forward pass of a surrogate model over the full library can be prohibitively expensive. 
This bottleneck violates a core assumption of standard active learning pipelines that surrogate inference is negligible relative to evaluation cost \citep{bo-tutorial, garnett_bayesoptbook_2023}.

This shift in the cost hierarchy reframes molecular discovery as a large-scale discrete decision-making problem under dual constraints on evaluation and inference. 
To address this challenge, we introduce \textsc{BoBa} (Bayesian Optimization with BAndits), which explicitly accounts for inference cost by combining structure-aware partitioning of the action space, bandit-based allocation across partitions, and surrogate-guided optimization within partitions. 
By decoupling global and local search, \textsc{BoBa} enables efficient candidate selection without exhaustive inference over the action space. 
Systematic benchmarks on real-world drug discovery data demonstrate the critical importance of a) bandit strategies which explicitly account for uncertainty, rather than relying randomized exploration, and b) a chemically sensible partitioning of action space. 
Our empirical results demonstrate a tunable tradeoff between optimization performance and inference cost, which allows the optimization efficiency of complete-inference BO to be largely retained, at substantially reduced inference cost. 
Scaling experiments up to approximately $10^8$ molecules show that this tradeoff becomes increasingly favorable as library size grows, and a simple theoretical analysis identifies the number of partitions as the key control parameter governing the balance between inference savings and bandit regret.
These results lay the foundation for scaling to virtual libraries that contain billions to trillions of candidates. 

\section{Preliminaries}

This section introduces the optimization setting considered in this work and reviews relevant concepts from Bayesian optimization and multi-armed bandits (MAB).

\subsection{Virtual Screening as Large-Scale Discrete Optimization}

We consider the problem of computational candidate selection in virtual screening. 
Let $\mathcal{X} = \{x_1, \dots, x_N\}$ denote a finite library of candidate molecules, where $N$ may range from millions to trillions. 
Each molecule $x \in \mathcal{X}$ is associated with an unknown property of interest $f(x) \in \mathbb{R}$, such as docking score, binding affinity, or experimental activity, which can only be accessed through an expensive evaluation (e.g., docking or wet-lab assay).

The goal of virtual screening is to efficiently identify molecules with high values of $f(x)$ using as few evaluations as possible, framing virtual screening as a discrete optimization problem,
\[
x^\ast = \argmax_{x \in \mathcal{X}} f(x),
\]
under a limited evaluation budget $T \ll |\mathcal{X}|$.

In addition to evaluation cost, we explicitly consider the computational cost of surrogate inference over $\mathcal{X}$. 
In modern make-on-demand settings, $|\mathcal{X}|$ is sufficiently large that exhaustive scoring of all candidates using a learned model is itself infeasible. 
We therefore distinguish between:
(i) \emph{evaluation cost}, incurred when querying $f(x)$, and 
(ii) \emph{inference cost}, incurred when computing surrogate predictions over subsets of $\mathcal{X}$.
This distinction is central to the problem setting addressed in this work.

\subsection{Surrogate Modeling and Bayesian Optimization}

Bayesian optimization (BO) addresses black-box optimization by maintaining a probabilistic surrogate model over $f$ \citep{garnett_bayesoptbook_2023}. 
Given a dataset $\mathcal{D}_t = \{(x_i, y_i)\}_{i=1}^t$ with $y_i = f(x_i) + \epsilon_i$, the surrogate defines a posterior predictive distribution
\[
p(f(x) \mid \mathcal{D}_t),
\]
which is used to construct an acquisition function $a_t(x)$ that balances exploration and exploitation.

In discrete settings, BO typically proceeds by selecting
\[
x_{t+1} = \arg\max_{x \in \mathcal{X}} a_t(x).
\]
Unless $|\mathcal{X}|$ is large, this step is commonly approximated by exhaustively evaluating $a_t(x)$ over $\mathcal{X}$. 
However, this assumption becomes invalid for ultra-large libraries, where surrogate inference over $\mathcal{X}$ constitutes a computational bottleneck. 
This computational barrier motivates algorithmic strategies that avoid full-library surrogate evaluations while retaining the benefits of Bayesian decision-making.

\subsection{Multi-Armed Bandits}

Multi-armed bandits (MAB) \citep{Robbins1952, Lattimore2020} formalize sequential decision-making under uncertainty when limited resources must be allocated among competing alternatives.
At each round $t$, an agent selects an arm $k \in \{1, \dots, K\}$ and observes a stochastic reward drawn from an unknown distribution associated with that arm.
Bandit algorithms adaptively trade off exploration and exploitation to identify high-reward arms or to maximize cumulative reward.

In this work, each arm corresponds to a subspace of chemical space, and the observed reward summarizes the utility of recent evaluations from that region.
For the following discussion of Bandit policies, let $\hat{\mu}_k$ be the empirical mean reward of arm $k$, and $n_k$ the number of times arm $k$ has been selected.

\paragraph{$\epsilon$-Greedy} selects the empirically best arm with probability $1-\epsilon$, and explores by selecting an arm uniformly at random with probability $\epsilon$.
Although simple and computationally inexpensive, $\epsilon$-greedy does not explicitly account for uncertainty, which can lead to inefficient allocation when the number of arms is large.

\paragraph{Softmax Sampling} selects arms stochastically in proportion to their empirical mean rewards,
\[
P(k_t = k) \propto \exp(\tau \hat{\mu}_{k,~t-1}),
\]
where $\tau > 0$ is an inverse-temperature parameter controlling how strongly the distribution concentrates on high-reward arms.
Softmax provides a smooth tradeoff between exploration and exploitation, interpolating between uniform sampling and greedy selection.

\paragraph{Upper Confidence Bound (UCB1)} selects arms optimistically based on both empirical performance and uncertainty \citep{Auer2002}.
At round $t$, UCB1 chooses the arm maximizing
\[
\hat{\mu}_{k,~t-1} + c \sqrt{\frac{2 \log t}{n_{k,~t-1}}},
\]
where $c > 0$ controls the exploration--exploitation tradeoff.

Formal definitions and implementation details are provided in Appendix~\ref{app:mab}.

\subsection{Partitioning Methods for Chemical Space}

Let $\mathcal{X} = \{x_1, \dots, x_N\}$ denote a virtual library of molecules $x_i$.
When partitioning such virtual libraries, we consider dividing $\mathcal{X}$ into $K$ disjoint subsets $\{\mathcal{X}_1, \dots, \mathcal{X}_K\}$.
Typically, such partitioning is performed in a molecular feature space induced by a representation $\phi(x) \in \mathbb{R}^d$, which can be either learned or engineered. 

\subsubsection{Molecular Features}

\paragraph{Topological features} encode molecular structure as graph-derived patterns capturing atom connectivity and substructures (e.g. paths, cycles, and local neighborhoods). They often provide sparse, discrete representations ("fingerprints") optimized for similarity search.

\paragraph{Physicochemical descriptors} are engineered low-dimensional features that summarize global molecular properties derived from the graph structure (e.g. molecular weight, hydrogen bond donor count, polarity, or solubility). They offer interpretable, physically meaningful signals, but may miss fine-grained structural detail. The full list of descriptors used in this work is provided in Appendix~\ref{descriptors}.

\paragraph{Foundation model embeddings} are dense, learned representations produced by deep neural networks pretrained on large molecular corpora using self-supervised objectives. Pre-trained networks can include graph neural networks and language models operating on SMILES, a string-based encoding of the molecular graph structure. In this work, we focus on embeddings from the T5Chem model \citep{christofidellis2023unifying}, a chemistry-specific variant of the T5 architecture \citep{2020t5} pretrained on large corpora of molecular structures and textual descriptions. Prior work has shown that these embeddings capture rich chemical diversity and improve performance in molecular property prediction and active learning compared to traditional fingerprints \cite{kristiadi2024sober}.

\subsubsection{Partitioning Techniques}

\paragraph{Feature-based stratification} deterministically partitions molecules using fixed intervals along a small number of hand-crafted features, yielding axis-aligned regions in feature space. This approach has previously been used to partition ultra-large chemical libraries into chemically coherent regions for virtual screening \citep{Gorgulla2023}.

\paragraph{$k$-Means Clustering} groups data by assigning points to the nearest of $k$ learned centroids in feature space, optimized by minimizing within-cluster variance. 
\[
\sum_{k=1}^K \sum_{x \in \mathcal{X}_k} \|\phi(x) - \mu_k\|_2^2,
\]

As a baseline, we construct unstructured subspaces by randomly permuting $\mathcal{X}$ and dividing it into $K$ equally sized bins.
This randomization removes all chemical structure while preserving cluster sizes, isolating the effect of meaningful space decomposition.

\section{Method}

Building on the preliminaries, we formalize the problem of surrogate-based discovery from ultra-large molecular libraries, and introduce the algorithmic framework of \textsc{BoBa}. We also summarize related works relevant to the method proposed herein. 

\subsection{Problem Setting}

Given a large discrete library $\mathcal{X}$, an unknown black-box function $f$ (e.g. a docking score), and and a budget $T$ on the number of function evaluations, we seek an algorithm that sequentially selects candidates $\mathcal{C} = \{x_1, \dots, x_T\} \subset \mathcal{X}$ under these constraints. 
At each time point $t$, we assume access to a molecular representation $\phi(x) \in \mathbb{R}^d$, and a parametric or nonparametric surrogate model trained on all previous observations $\mathcal{D}_t = \{(x_i, f(x_i))\}_{i=1}^t$. 
Let $\mathcal{T}_m \subset \mathcal{X}$ denote the set of elements corresponding to the largest values of $f$.

In this scenario, a design strategy should select candidates such that
\begin{itemize}
\item[a)]{the number of top-$m$ elements recovered (i.e., $|~\mathcal{C} \cap \mathcal{T}_m~|$) is maximized.}
\item[b)]{the number of surrogate inferences is minimized.}
\end{itemize}


\subsection{Formulation of \textsc{BoBa}}

\textsc{BoBa} integrates surrogate-based optimization with structured allocation of inference and evaluation budgets. The fundamental concept entails decomposing $\mathcal{X}$ into $K$ subsets $\{\mathcal{X}_1, \dots, \mathcal{X}_K\}$, each of which serves as an arm in a multi-armed bandit problem.

At each iteration, a bandit algorithm selects a subset $\mathcal{X}_k$ based on past observations. 
Pulling arm $k$ triggers a localized optimization procedure within $\mathcal{X}_k$: surrogate inference is performed only on this subset, an acquisition function is evaluated locally, and a small batch of $B$ candidates is selected for expensive evaluation. 
The newly observed data are then added to $\mathcal{D}_t$, and the surrogate is updated. Ultimately, the observed reward becomes an aggregate metric of the utility of recent evaluations from that subspace (e.g., average docking score) and is used to guide future allocation (see Algorithm~\ref{alg:boba_main}).

\begin{algorithm}[h]
\caption{\textsc{BoBa}: Bayesian Optimization with Bandits over Clustered Subspaces}
\label{alg:boba_main}
\begin{algorithmic}[1]
\REQUIRE Library $\mathcal{X}$; expensive oracle $f(\cdot)$; featurizer $\phi(\cdot)$; number of subspaces $K$; bandit algorithm $\mathcal{B}$; surrogate model class $\mathcal{M}$; acquisition function $a(\cdot)$; rounds $T$; batch size $B$
\STATE \textbf{Cluster} $\mathcal{X}$ into $K$ disjoint subspaces $\{\mathcal{X}_1,\dots,\mathcal{X}_K\}$ using features $\phi(x)$
\STATE Initialize dataset $\mathcal{D}_0 \leftarrow \emptyset$ (or a small seed set)
\STATE Initialize bandit state for arms $k \in \{1,\dots,K\}$ in $\mathcal{B}$
\FOR{$t = 1,\dots,T$}
    \STATE Select an arm (subspace) $k_t \leftarrow \mathcal{B}(\text{history})$
    \STATE Fit/update a \textbf{global} surrogate $M_t \in \mathcal{M}$ on $\mathcal{D}_{t-1}$
    \STATE \textbf{Localized inference:} compute acquisition scores $\{a_t(x)\}_{x \in \mathcal{X}_{k_t}}$ using $M_t$
    \STATE Select a batch:\\$\mathcal{Q}_t \leftarrow \text{Top-}B\ \text{elements of }\mathcal{X}_{k_t}\text{ by }a_t(x)$
    \STATE Evaluate oracle: obtain $y_t(x) \leftarrow f(x)$ for all $x \in \mathcal{Q}_t$
    \STATE Update dataset: $\mathcal{D}_t \leftarrow \mathcal{D}_{t-1} \cup \{(x, y_t(x)) : x \in \mathcal{Q}_t\}$
    \STATE Compute arm reward (batch average):\\$r_t \leftarrow \frac{1}{B}\sum_{x \in \mathcal{Q}_t} y_t(x)$
    \STATE Update bandit state: $\mathcal{B} \leftarrow \textsc{Update}(\mathcal{B}, k_t, r_t)$
\ENDFOR
\STATE \textbf{Output:} $\arg\max_{(x,y)\in \mathcal{D}_T} (y)$
\end{algorithmic}
\end{algorithm}

This perspective reframes large-library screening as a hierarchical decision problem:
rather than repeatedly scoring all candidates, the algorithm first decides \emph{where} in chemical space to invest computational resources, and only then performs fine-grained surrogate-based selection within the chosen region.
When subspaces differ in their intrinsic concentration of high-quality molecules, bandit algorithms provide a principled mechanism for identifying and exploiting promising regions while continuing to explore uncertain ones.
This formulation is central to \textsc{BoBa}, enabling adaptive, data-driven control over inference and evaluation budgets without requiring exhaustive scans of the action space.

Within this work we perform extensive experimental benchmarks of \textsc{BoBa}: 
First, we evaluate different bandit algorithms and analyze their impact on optimization performance and sample efficiency. 
Second, we systematically study the tradeoff between optimization performance and computational cost by varying the number and size of partitions, highlighting how granularity affects inference cost. 
Third, we examine how different choices of featurization and partitioning strategy influence optimization performance, comparing structured clustering approaches to random partitions.

\begin{figure*}[t]
    \centering
    \begin{subfigure}{0.48\linewidth}
        \centering
        \includegraphics[width=\linewidth]{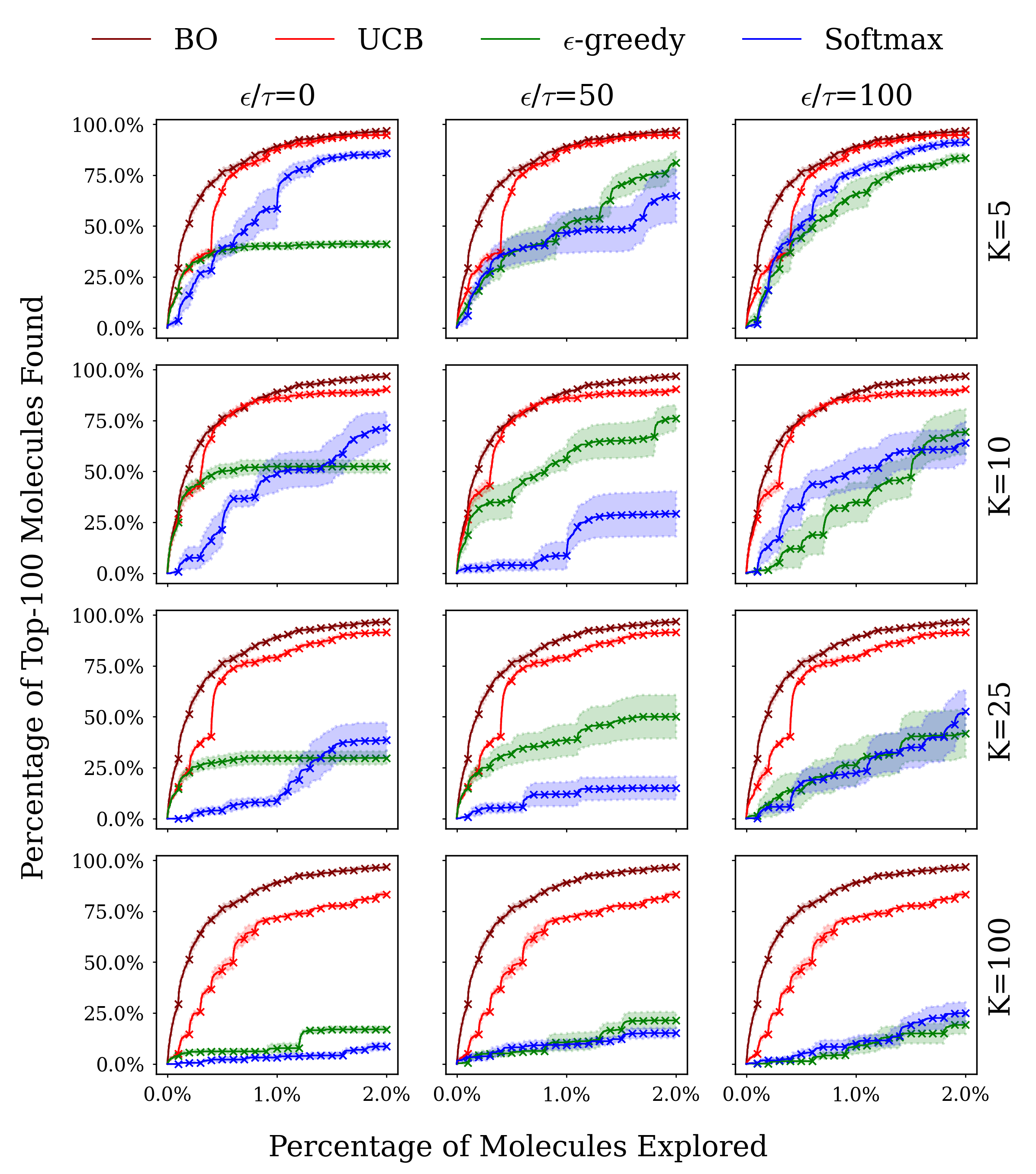}
        \label{fig:boba_a}
    \end{subfigure}
    \hfill
    \begin{subfigure}{0.48\linewidth}
        \centering
        \includegraphics[width=\linewidth]{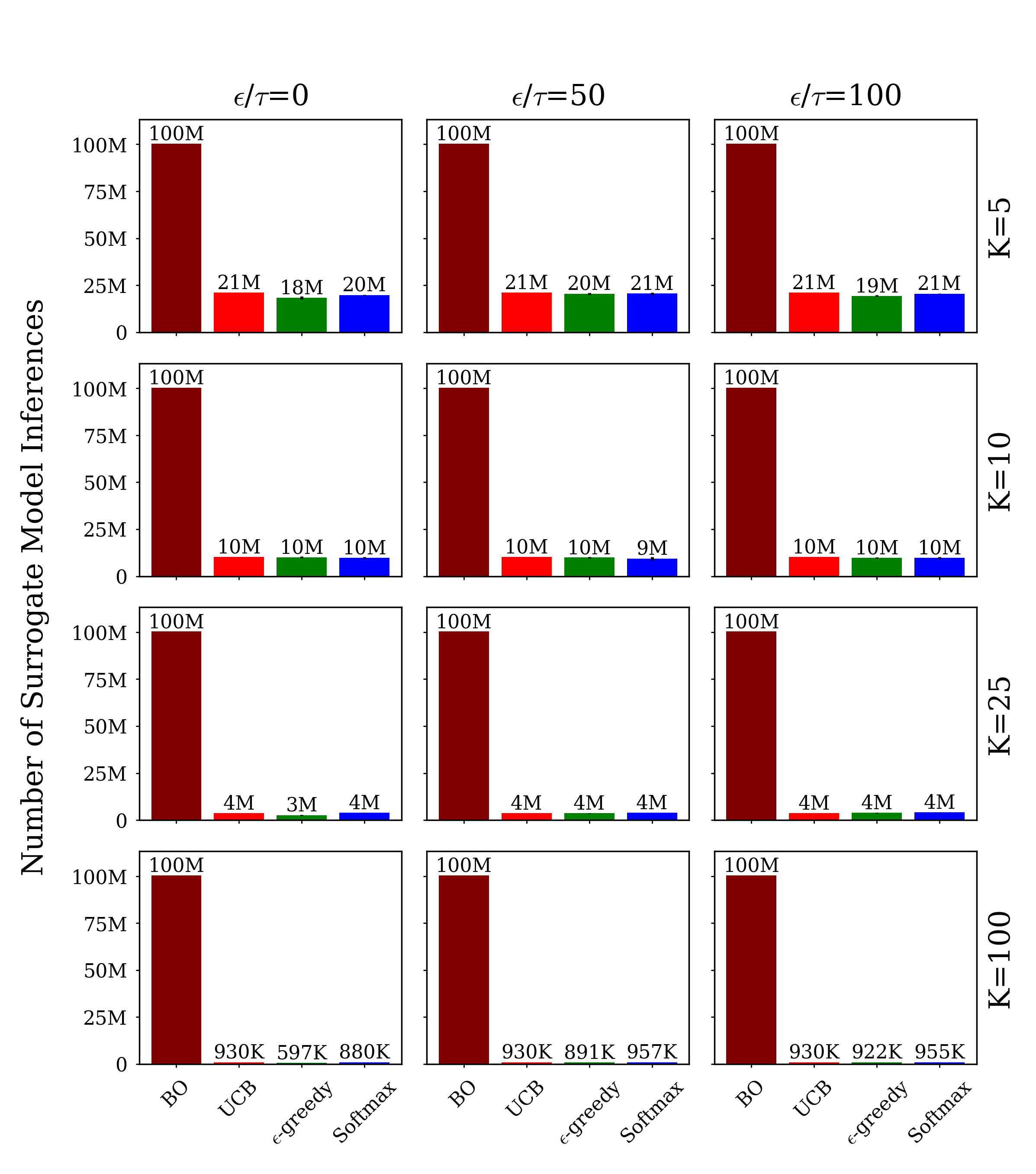}
        \label{fig:boba_b}
    \end{subfigure}
    \caption{\textbf{Evaluation of Bandit algorithms, and observed tradeoff between performance and inference cost}.
    (\textbf{Left}) Optimization trajectories of \textsc{BoBa} with different Bandit algorithms as a function of the number of costly black-box evaluations. BO with full-library inference (maroon) is included as an upper boundary. Performance is quantified by the number of retrieved molecules from the top-100 candidates from the full library.
    (\textbf{Right}) Computational cost of the respective optimization runs, measured by the number of surrogate model inferences. 
    All experiments are reported on the Enamine-5M library docked against CKB. Trajectories are shown as the mean over 5 independent runs from different seed populations. The shaded area indicates the standard error of the mean. 
    }
    \label{fig:boba_bandits}
\end{figure*}

\subsection{Related Work}

Active learning strategies are widely used in large-library virtual screening and have been reviewed extensively elsewhere.\cite{Reker2015}
Here, we focus on prior work that combines active learning with explicit segmentation of large molecular libraries. 
In general, virtual libraries are constructed by combinatorial enumeration from sets of synthetic precursors (\textit{synthons}), and several approaches exploit this structure to improve search efficiency. 
A common strategy is to perform search within this low-dimensional discrete synthon space, which avoids explicit enumeration of the full library. This setting has been addressed using active learning approaches \citep{Grigg2025, Kozyrev2025}, as well as multi-armed bandit formulations \citep{Klarich2024, Zhao2025}. 
Relatedly, the synthon-based structure of virtual libraries has also been exploited for hierarchical coarse-to-fine retrieval of candidate molecules \citep{Nazarova2025}. 

Our work differs both in how structure is imposed on the search space and how this structure is exploited algorithmically. 
Existing approaches typically operate over a fixed synthon-defined hierarchical organization of the library. 
In contrast, \textsc{BoBa} uses explicit molecular-structure-aware partitioning, and combines it with bandit-based methods for dynamic allocation of sampling effort across partitions, and surrogate-based optimization for efficient candidate selection within each partition. 
This combination allows the algorithm to adaptively balance exploration and exploitation across heterogeneous regions of the library, rather than committing to a predefined partitioning or a single level of resolution.
The resulting search strategy uses target feedback to allocate effort across learned molecular partitions, without requiring the fixed synthon-level hierarchy assumed by fragment-based search methods.

\section{Benchmarking Experiments}

We benchmark \textsc{BoBa} on virtual libraries derived from ultra-large chemical spaces, as complete ground-truth docking scores are available for these libraries. 
Specifically, we randomly selected 5 million compounds from an enumerated library of approximately 69 billion molecules from \textsc{Enamine REAL} \citep{enamine, Gorgulla2023}, which we refer to as \textsc{Enamine-5M}. 
In addition, we sampled an independent set of 3.9 million compounds from \textsc{Enamine}’s S-class small-molecule database \citep{enamine}, which we refer to as \textsc{Enamine-S-3.9M}.
Both libraries were exhaustively docked against the protein targets NEDD4 \citep[PDB: 9HT9;][]{Maspero2025}, and CKB \citep[PDB: 3B6R;][]{Bong2008}. 
In addition, we use a virtual library of 2M molecules from \textsc{Enamine}'s HTS database (\textsc{Enamine-HTS}), which was docked against a Thymidylate Kinase (TMK) by \citep{Graff2021}.

\textsc{BoBa} uses a feedforward neural network surrogate operating on fixed molecular features, with uncertainties approximated \textit{via} a Linearized Laplace approximation \cite{laplace2021} (see Appendix~\ref{app:bo_details} for more details). 
We additionally evaluate alternative uncertainty estimators, including MC dropout \citep{Gal2016Dropout} and SWAG \citep{Maddox2019SWAG} (see Appendix~\ref{appendix:uncertainty} and Figure~\ref{fig:uncertainty_ablation}); these experiments support the use of the Laplace approximation in all further experiments.

After a single-round initialization with randomly drawn samples from all partitions, \textsc{BoBa} follows the algorithm outlined in Algorithm~\ref{alg:boba_main}.
From a bandit-selected partition, all molecules are scored by the neural network surrogate, and a batch of $B$ candidates are selected using the Upper Confidence Bound as the acquisition function. 
Across all methods, we run $T=20$ rounds of selection, with a batch size of $B=5{,}000$ for a total of $TB = 100{,}000$ oracle evaluations. In the main text, we report optimization trajectories for the number of top-100 molecules from the full library that were found by the algorithm, which is motivated by budget constraints on downstream experimental validation. 
Trajectories for the top-1000 and top-10000 molecules are provided in the Appendix~\ref{appendix:results}, specifically Figures~\ref{fig:ckb5m-topK},~\ref{fig:ckb3.9m-topK}, and~\ref{fig:nedd3.9m-topK}.

\subsection{Bandit Exploration Strongly Impacts Screening Performance}

We first examine how the choice of bandit algorithm affects optimization performance. 
Specifically, we evaluate $\epsilon$-greedy, softmax sampling, and UCB1 bandits across different levels of exploration, and under variation of the granularity of library partitions. 
The resulting optimization trajectories and inference cost estimates on the Enamine-5M library, docked against CKB, are shown in Figure~\ref{fig:boba_bandits}. 
As an upper-bound estimate of optimization performance, we compare against standard BO with full-library inference. 

Empirically, we find that $\epsilon$-greedy and softmax exhibit substantial degradation of optimization performance as the number of library partitions increases.
We attribute this behavior to the lack of principled uncertainty-guided exploration in these methods, which can lead to premature over-exploitation of suboptimal partitions, or insufficient exploitation of optimal ones. 
UCB1 instead assigns an optimism bonus to under-sampled arms, so uncertain partitions remain eligible for selection even when their empirical rewards are initially modest.
Consistent with this mechanism, the per-arm selection frequencies in Appendix~\ref{appendix:arm-selection}, specifically Figure~\ref{fig:arm-selection}, show that UCB1 continues to allocate evaluations across multiple partitions throughout optimization, without collapsing onto a single early winner.
Thus, UCB1 remains largely competitive with full-inference BO even at large numbers of partitions, recovering nearly the same number of top-100 ranked molecules at a substantially reduced inference cost. 
Notably, the trends remain unchanged when considering the top-1000 and top-10000 candidate molecules from the full library (see Appendix~\ref {appendix:results}, specifically Figures~\ref{fig:ckb5m-topK},~\ref{fig:ckb3.9m-topK}, and~\ref{fig:nedd3.9m-topK}).

These results indicate that, particularly in regimes with many arms, optimism-based allocation is critical, which motivates the use of UCB1 in all subsequent experimental studies. 
Remarkably, even this relatively simple algorithm achieves strong performance, suggesting that more advanced uncertainty-aware bandit methods may offer additional benefits.
However, their analysis is beyond the scope of this study.

Finally, even for the UCB1 algorithm, we observe a tradeoff between optimization performance and inference cost. 
This trade-off can be tuned by the user through the choice of partition granularity (that is, the number of clusters), depending on the requirements of a specific use case.
The role of $K$ can be understood through a simple cost--regret decomposition.
Partitioning reduces the cumulative surrogate inference cost from order $NT$ to approximately $NT/K$, but increasing $K$ also makes the bandit allocation problem harder because the algorithm must identify promising regions among more arms.
Under a gap-free UCB-style regret term of order $\tilde{\mathcal{O}}(\sqrt{KT})$, minimizing a weighted objective of inference cost and allocation regret yields the heuristic scaling for choosing $K$ as $K^\ast \asymp N^{2/3}T^{1/3}$.
This analysis, detailed in Appendix~\ref{appendix:theory}, formalizes why larger libraries can support finer partitions while still requiring a sublinear choice of $K$.

\subsection{Structured Subspaces are Critical}

We next perform a systematic ablation to understand whether the gains of \textsc{BoBa} arise merely from partitioning the library, a persistent partitioning scheme, or from the inherent library structure captured by clustering.

\subsubsection{Static \textsc{BoBa} vs.\ Dynamic Partitioning}

First, we compare \textsc{BoBa} to an unstructured baseline in which, at each iteration, standard BO is applied to a uniformly random subset of the library whose size matches the average cluster size in \textsc{BoBa}.
Related subsampling strategies have been used in prior works to mitigate the cost of exhaustive surrogate inference \citep{wang-henderson2023graph}.

\begin{figure}
    \centering
    \includegraphics[width=\linewidth]{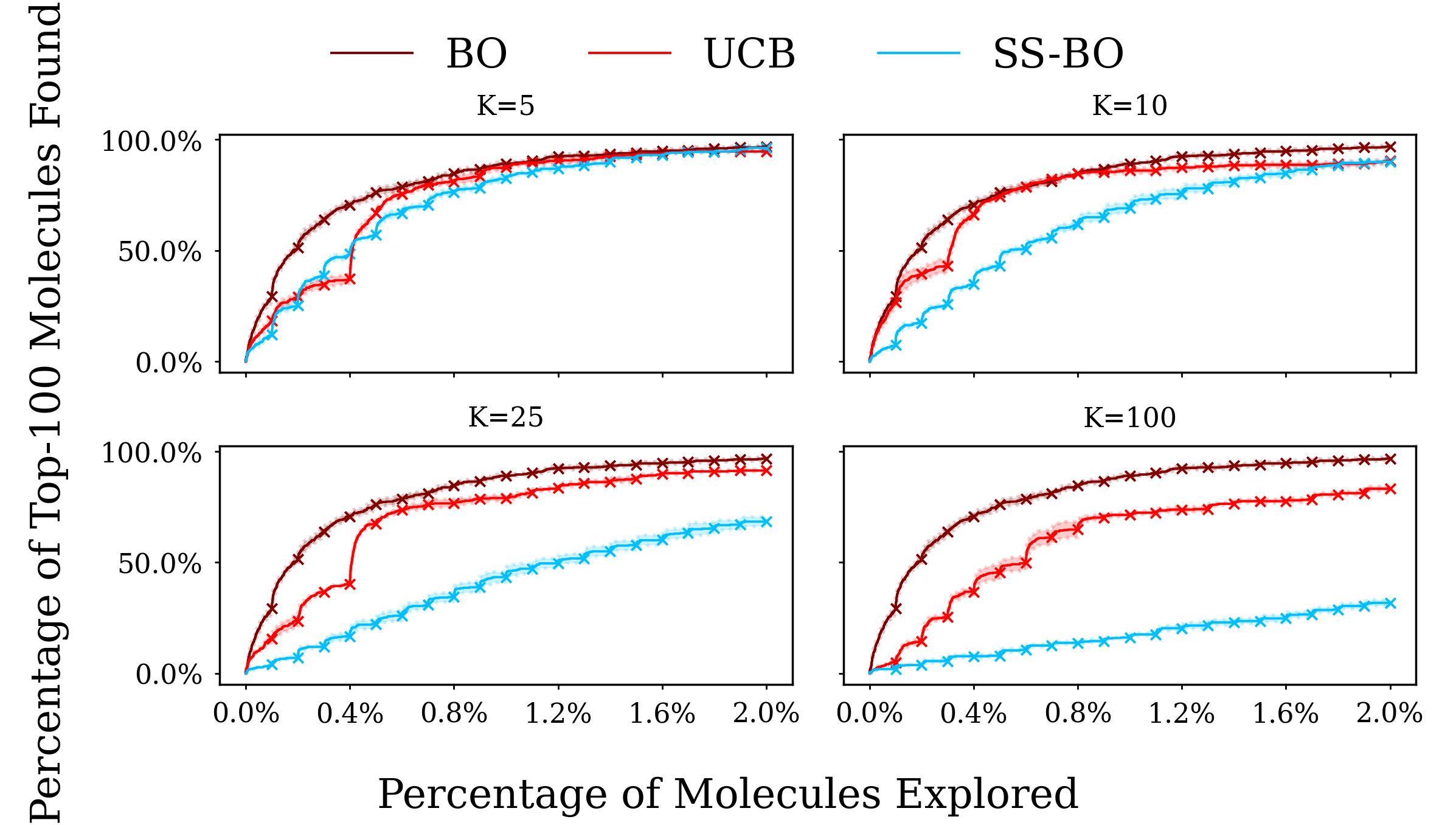}
    \caption{\textbf{Influence of static vs. dynamic partitioning.}
    Optimization trajectories of \textsc{BoBa} with UCB1 using $K$-Means-based partitions, compared to setting in which a partition for inference is randomly selected at each iteration. 
    All experiments are reported on the Enamine-5M library docked against CKB. Trajectories are shown as the mean over 5 independent runs from different seed populations. 
    The shaded area indicates the standard error of the mean.}
    \label{fig:boba_subsample}
\end{figure}

Figure~\ref{fig:boba_subsample} shows that \textsc{BoBa} substantially outperforms random subsampling across all settings.
While both methods reduce inference cost to the same degree, subsampling repeatedly reallocates computation to arbitrary regions of chemical space.
In contrast, \textsc{BoBa} defines persistent partitions and uses bandit feedback to adaptively concentrate evaluations in empirically promising regions.
These findings indicate that \textsc{BoBa} achieves significant optimization performance gains by identifying persistent partitions of the action space, and adaptively allocating resources to these partitions. 

\subsubsection{Structured vs.\ Randomly Assigned Partitions}

Second, we investigate whether a static library partitioning alone is sufficient to maintain competitive optimization performance at reduced inference budget. Therefore, we compare \textsc{BoBa} using $K$-means-based molecular clusters to a variant in which molecules are randomly assigned once to $K$ equally sized, static subsets, which are then treated as arms by the same UCB1 algorithm. 
Both approaches use identical surrogates, acquisition functions, batch sizes, and bandit rules; they differ only in how the partitions are initially defined.

\begin{figure}[h]
    \centering
    \includegraphics[width=\linewidth]{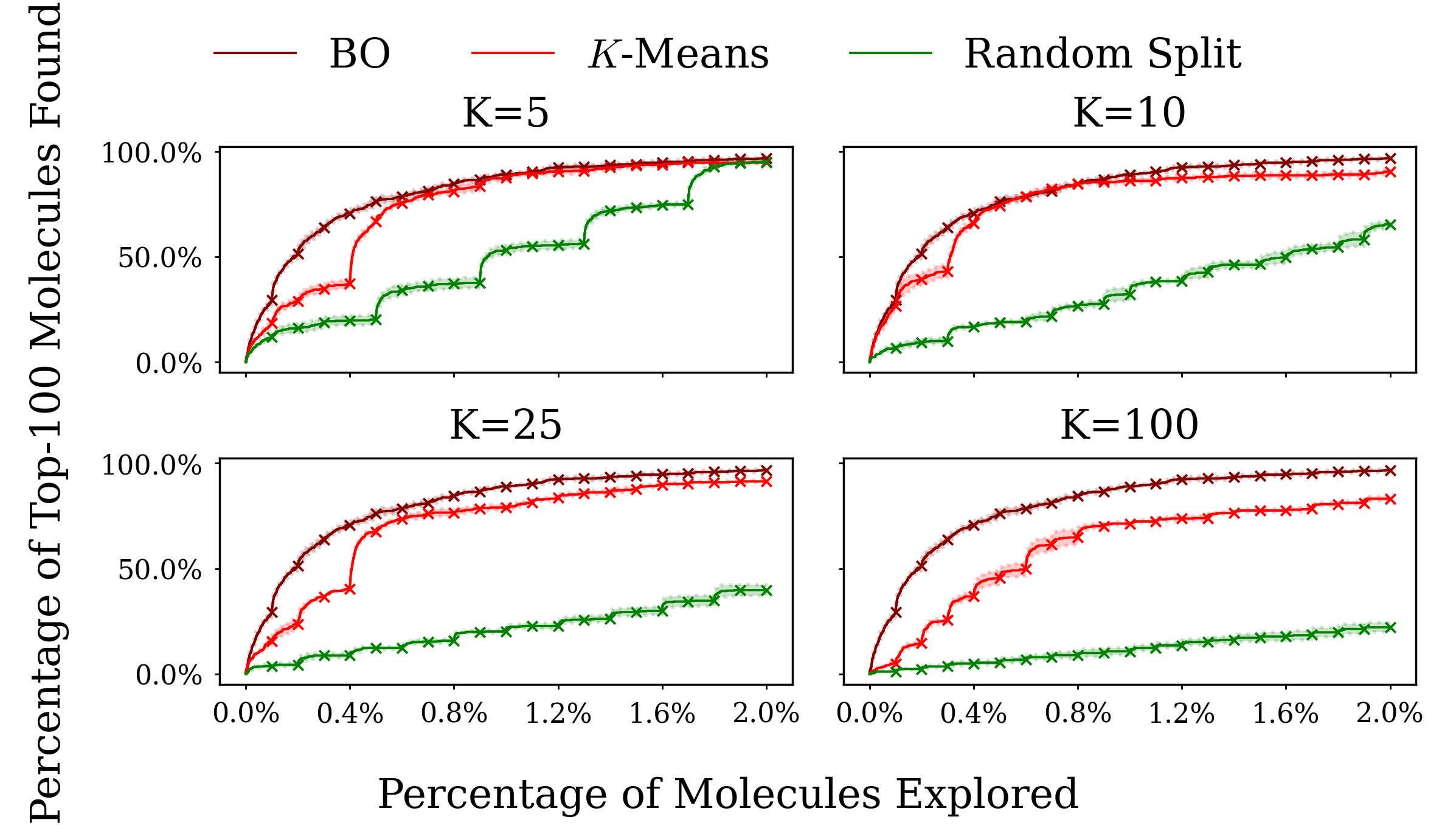}
    \caption{\textbf{Influence of the partitioning scheme.}
    Optimization trajectories of \textsc{BoBa} with UCB1 using either $K$-Means-based partitions, or randomly assigned partitions.
    All experiments are reported on the Enamine-5M library docked against CKB. Trajectories are shown as the mean over 5 independent runs from different seed populations. 
    The shaded area indicates the standard error of the mean.}
    \label{fig:boba_splitk}
\end{figure}

As shown in Figure~\ref{fig:boba_splitk}, replacing the structured partitions obtained \textit{via} $K$-means clustering with random partitions leads to a substantial degradation in optimization performance.
Although both methods maintain persistent partitions and operate under identical inference budgets, only structured clustering yields arms that are systematically enriched for high-quality candidates.
Analysis of the distribution of targets across partitions (see Appendix~\ref{appendix:score-dist}, specifically Figures~\ref{fig:scores_ckb-5m},~\ref{fig:scores_ckb-3.9m}, and~\ref{fig:scores_nedd4-3.9m}, for further details) confirms the presence of significant inter-partition variance in the case of clustering, which is a prerequisite for effective exploration–exploitation across arms. 
These findings confirm that the effectiveness of \textsc{BoBa} depends critically on the chemical coherence of the partitions, rather than on a static partition of the library alone.

\subsection{Effect of Feature Space}

Having established the necessity of structured partitioning for achieving competitive optimization performance at reduced inference budget, we next evaluate how the molecular representation used to construct partitions impacts end-to-end \textsc{BoBa} performance. 
Therefore, we compare a feature space spanned by established physicochemical descriptors for drug discovery \cite{Gorgulla2020} without further refinement, with the embedding space of a domain-specific language model (T5Chem). 

\begin{figure}[h]
    \centering
    \includegraphics[width=\linewidth]{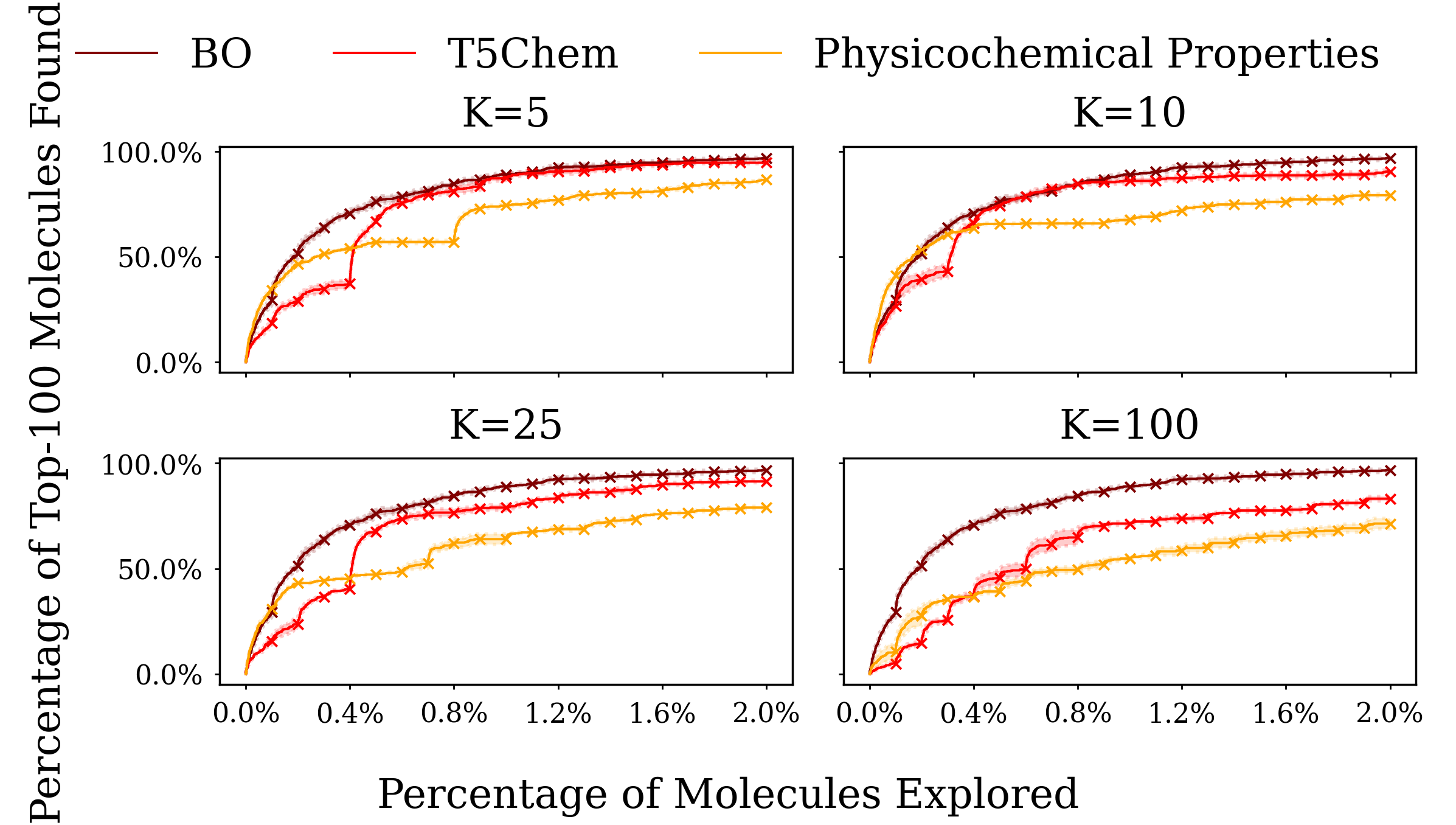}
    \caption{\textbf{Effect of feature space on end-to-end \textsc{BoBa} performance.}
    Optimization trajectories of \textsc{BoBa}, with partitions constructed using T5Chem language model embeddings, or using unrefined physicochemical descriptors.
    All experiments are reported on the Enamine-5M library docked against CKB. Trajectories are shown as the mean over 5 independent runs from different seed populations. 
    The shaded area indicates the standard error of the mean.}
    \label{fig:boba_features}
\end{figure}

Figure~\ref{fig:boba_features} demonstrates that, across all values of $K$, \textsc{BoBa} constructed in T5Chem embedding space consistently outperforms \textsc{BoBa} constructed in unrefined physicochemical descriptor space, recovering substantially more top-ranked molecules under identical budgets.
These findings suggest that clustering in T5Chem embedding space produces partitions that are more closely aligned with the optimization objective, allowing the bandit to allocate more effectively across regions, while improving local surrogate generalization.
In all experiments, the same molecular representation is used for both clustering and surrogate modeling.
This design choice ensures that the geometry used to define bandit arms is consistent with the geometry used for local acquisition, avoiding discrepancies between global partitioning and local optimization.
We therefore consider this the preferred default when the representation is sufficiently informative for the target task.
At the same time, this design also implies that biases from a poorly aligned representation are propagated into both stages, motivating the partion-quality ablations discussed above. 
These conclusions align well with the findings by \citep{kristiadi2024sober}, who reported that, even without task-specific fine-tuning, such embeddings provide an effective molecular representation across different tasks. 

Overall, these results indicate that the effectiveness of \textsc{BoBa} is tightly coupled to the representational geometry of chemical space.
While pre-trained molecular foundation models provide a robust starting point, we anticipate that integrating domain expertise into task-specific feature spaces can further improve or accelerate optimization with \textsc{BoBa}.

\subsection{Robustness Across Targets and Difficulty Regimes}

\begin{figure}[H]
    \centering
    \vspace{0.25cm}
    \begin{subfigure}[t]{\linewidth}
        \centering
        \textbf{\textsc{Enamine-S-3.9M} (CKB)}
        \includegraphics[width=\linewidth]{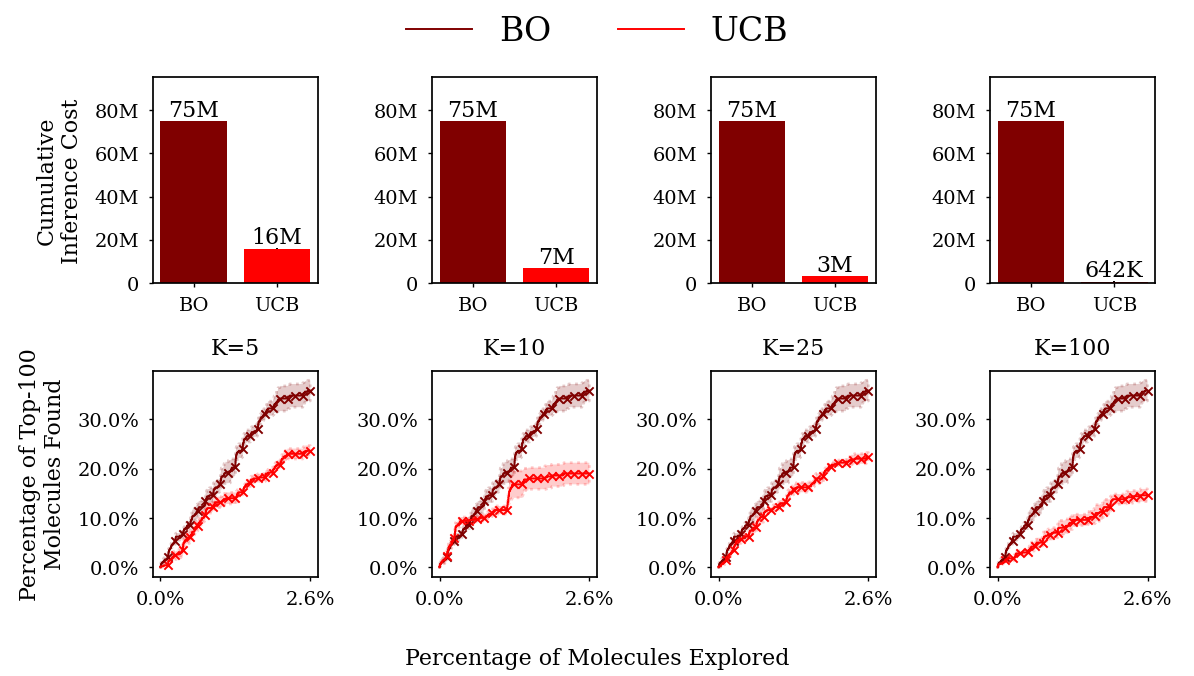}
        \label{fig:ckb_39m}
    \end{subfigure}

    \vspace{0.25cm}

    \begin{subfigure}[t]{\linewidth}
        \centering
        \textbf{\textsc{Enamine-S-3.9M} (NEDD4)}
        \includegraphics[width=\linewidth]{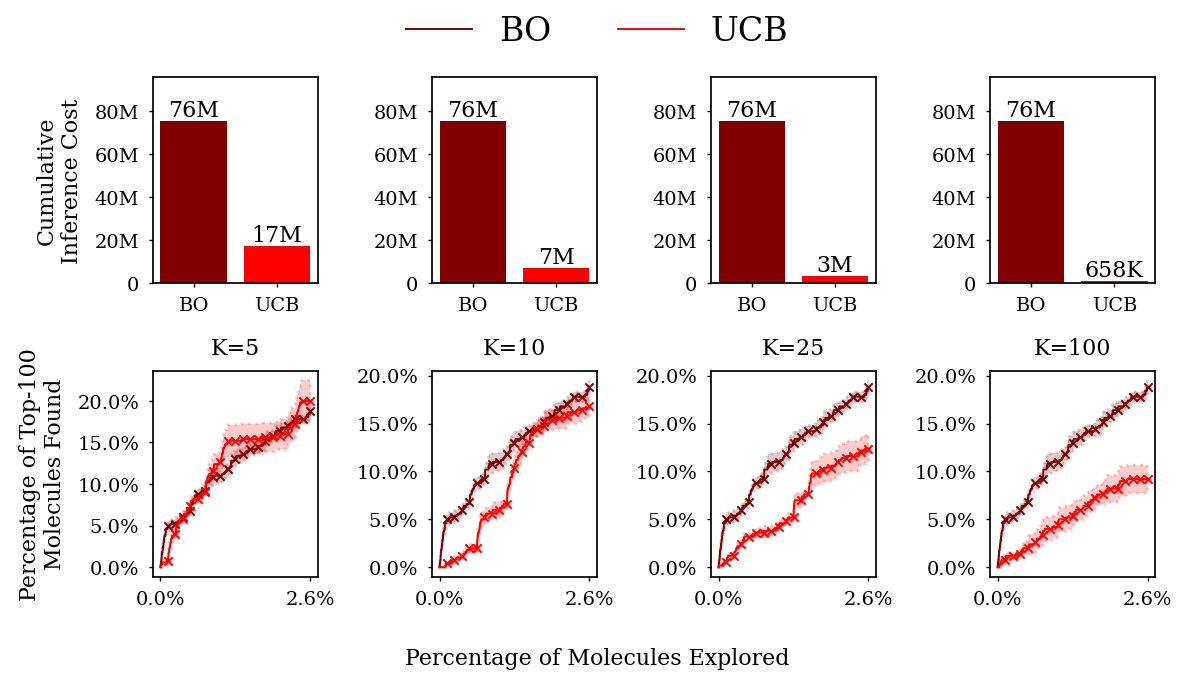}
        \label{fig:nedd4_39m}
    \end{subfigure}

    \vspace{0.25cm}

    \begin{subfigure}[t]{\linewidth}
        \centering
        \textbf{\textsc{Enamine-HTS} (TMK)}
        \includegraphics[width=\linewidth]{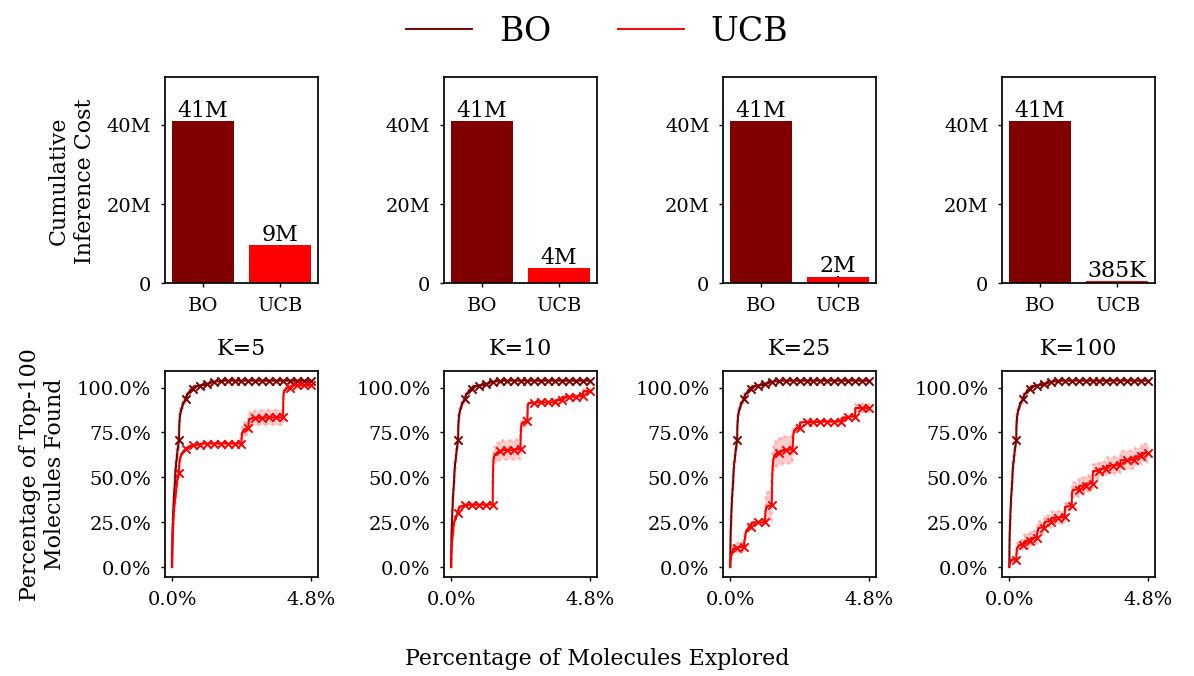}
        \label{fig:tmpk_hts}
    \end{subfigure}

    \caption{\textbf{\textsc{BoBa} across targets and difficulty regimes.}
    Optimization trajectories of \textsc{BoBa} on different optimization problems. 
    Trajectories are shown as the mean over 5 independent runs from different seed populations. 
    The shaded area indicates the standard error of the mean.
    }
    \label{fig:39m_targets}
\end{figure}

Finally, we study the optimization behavior of \textsc{BoBa} with UCB1 on three additional tasks: the \textsc{Enamine-HTS} library docked against TMK \citep{Graff2021}, and the \textsc{Enamine-S-3.9M} library docked against CKB and NEDD4. The corresponding optimization trajectories, compared to BO with full-library inference as an upper-bound estimate, are shown in Figure~\ref{fig:39m_targets}.

Qualitatively, we find that the trends discussed in the previous sections are reproduced across tasks. 
As $K$ increases, we observe a similar sublinear decay of optimization performance, indicating a feasible tradeoff between optimization performance and inference cost. 
Notably, the absolute optimization scores on the \textsc{Enamine-S-3.9M} library are substantially lower than those on the \textsc{Enamine-5M} (CKB) benchmark, both for the upper-bound estimate and, by extension, for \textsc{BoBa}.
This performance gap suggests that these tasks pose more challenging optimization problems. 
Indeed, analysis of the underlying library revealed a larger chemical diversity in the \textsc{Enamine-S-3.9M} library compared to the \textsc{Enamine-5M} library, which provides a possible explanation for the observed differences in absolute performance.

These results demonstrate that \textsc{BoBa}'s performance--inference tradeoff and dependence on cluster granularity persist across targets and difficulty regimes, indicating that the framework generalizes beyond a single dataset or optimization landscape.

\subsection{Scaling to Larger Libraries}

The preceding experiments evaluate \textsc{BoBa} in settings where exhaustive surrogate inference is still possible, which enables direct comparison against full-library BO. 
To test whether the observed tradeoff remains relevant at larger scale, we further evaluate \textsc{BoBa} on the ZINC library docked against AmpC \citep{Lyu2019} using nested libraries of approximately $10^5$, $10^6$, $10^7$, and $10^8$ molecules. 
For each library size, we compare the area under the top-1000 retrieval curve for \textsc{BoBa} against the corresponding full-library BO run, normalizing the BO performance to one.

\begin{figure}[t]
    \centering
    \includegraphics[width=\linewidth]{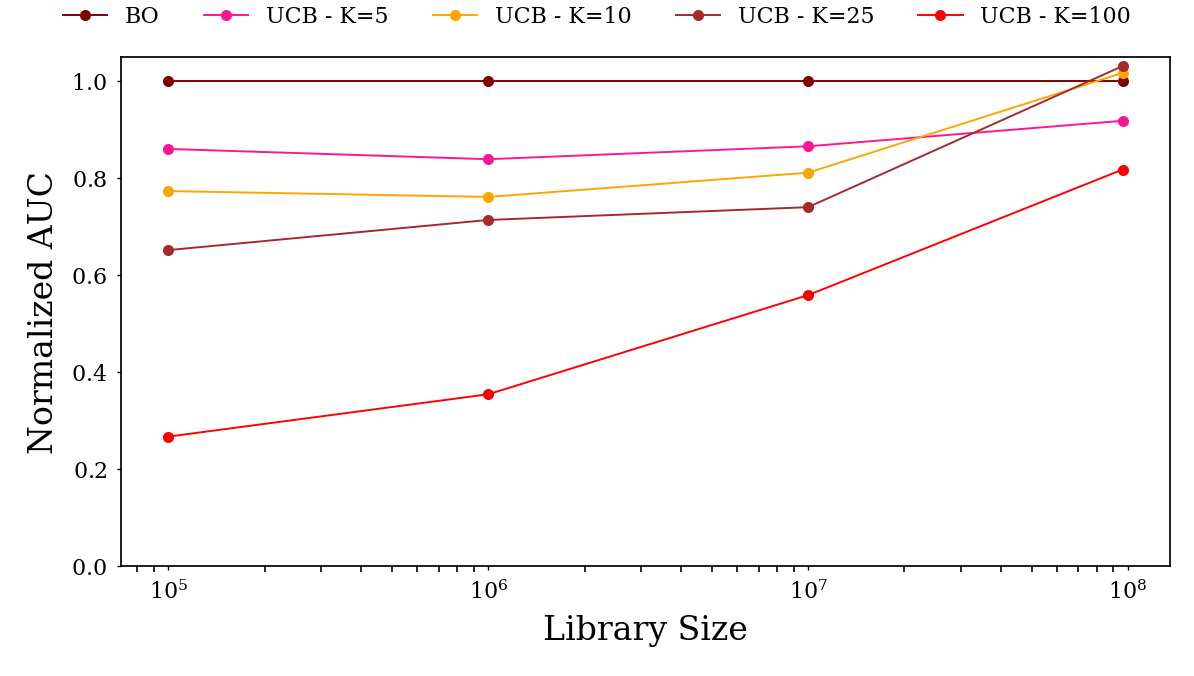}
    \caption{\textbf{Scalability of \textsc{BoBa} with library size.}
    Optimization performance of \textsc{BoBa} relative to full-library BO as the library size increases from approximately $10^5$ to $10^8$ molecules. 
    Performance is measured as the area under the top-1000 retrieval curve and normalized by the AUC of the corresponding full-library BO run.
    All experiments use the ZINC library docked against AmpC, or subsets thereof.}
    \label{fig:boba_scaling}
\end{figure}

As shown in Figure~\ref{fig:boba_scaling}, \textsc{BoBa}'s relative performance does not deteriorate as the candidate library grows.
Instead, for several partition granularities, the normalized AUC approaches full-library BO at the largest tested scale.
This trend is consistent with the core motivation of \textsc{BoBa}: as $N$ increases, full-library inference becomes increasingly costly, while localized inference over a selected partition remains controlled by the cluster granularity.
The value of $K$ therefore acts as a user-facing knob that trades inference cost against the statistical difficulty of selecting among more arms.
Appendix~\ref{appendix:runtime} reports the corresponding wall-clock breakdown.

\section{Conclusion and Outlook}

Ultra-large make-on-demand libraries push surrogate-based optimization into a regime where surrogate inference and acquisition over the full candidate set can become as limiting as the expensive oracle itself. 
In this work, we introduce \textsc{BoBa}, a target-aware framework that avoids full-library inference by partitioning chemical space into persistent subspaces, and using a multi-armed bandit to allocate inference and evaluation to regions that empirically yield high-utility molecules. 

Benchmark experiments highlight the critical influence of both the bandit exploration strategy and the structure of the partitioning scheme. 
Optimism-under-uncertainty bandits combined with clustering on foundation model embeddings consistently delivered robust performance across tasks, clearly outperforming clustering on unrefined physicochemical descriptors as well as unstructured baselines on randomized partitions. 
Looking forward, we anticipate that advances in bandit algorithms, particularly improved uncertainty quantification and rotting bandits to accommodate non-stationary partition rewards \citep{Levine2017}, will further enhance optimization performance. 
Coupling these methods with problem-specific, expert-refined molecular representations is a promising direction for further gains. 

Our findings further reveal a clear trade-off between optimization performance and inference cost. 
Empirical results across increasingly large libraries indicate that the decay in optimization performance with increasing numbers of partitions is sublinear, making it feasible for practitioners to select the number of partitions, and therefore the inference cost, according to problem-specific constraints. 
This behavior is supported by a simplified cost-regret analysis, which identifies the number of clusters as a natural control parameter and predicts an optimal granularity that grows sublinearly with library size.
While partitioning very large libraries introduces additional computational cost, the employed algorithm scales as $\mathcal{O}(nkid)$, dominated by the library size $n$, and constitutes a one-time processing expense that can be reused across targets. 
Assessing how broadly these results generalize across even larger libraries and more diverse targets remains the focus of ongoing work. 
Overall, this study represents a step toward principled decision-making at billion- to trillion-scale, in chemical discovery and other large discrete search problems, where both evaluation and inference are costly.

\section{Limitations}

The current implementation and empirical evaluation of \textsc{BoBa} are restricted to settings in which the virtual library is explicitly enumerable, and in which featurization and clustering of the full library remain computationally tractable. 
Conceptually, however, the proposed framework does not rely on explicit enumeration of candidate molecules. 
The combination of library partitioning, bandit-driven allocation of sampling effort across partitions, and surrogate modeling for candidate selection could instead be applied directly in synthon space. 
This extension would avoid the need for full library enumeration and global featurization.
Instead, enumeration, featurization, and inference would be required only for global subsamples or small, tractable partitions. 
We leave the implementation and systematic benchmarking of this synthon-level formulation to future work.

\pagebreak
\section*{Acknowledgements}

The authors thank Dr. Michael Emanuel for helpful discussions and Dr. Christoph Gorgulla for providing the Enamine-5M library. 
Y.C. acknowledges support from the BCMP Scholars Summer Undergraduate Research Program at Harvard Medical School.
C.K. was supported by a Hertz Foundation Fellowship and previously by NIH/NIGMS Molecular Biophysics Training Grant T32 GM008313. 
F.S.-K. acknowledges funding from the Deutsche Forschungsgemeinschaft (DFG) under the Priority Program 2363 "Molecular Machine Learning" (grant no. 497260357).
H.A. acknowledges support from Pivotal Life Sciences and NIH grant R35 GM158220 (NIGMS). 

\section*{Impact Statement}

This work aims to accelerate the identification of promising candidates from ultra-large discrete libraries, motivated by applications in early-stage drug discovery. 
By improving the efficiency of computational candidate selection, such methods may reduce the cost and time required to identify therapeutic leads, with potential downstream benefits for global healthcare.
The proposed framework addresses a general class of large-scale discrete optimization problems, and may therefore be applicable beyond drug discovery, e.~g. in materials design, catalyst discovery, or related scientific settings.
The societal consequences of such applications depend on the specific use case, but could be relevant to areas like energy storage and conversion, sustainable materials development, or improved resource and materials cycles.

At the same time, more efficient exploration of chemical space raises potential dual-use concerns. 
While the present approach is limited to computational prioritization and requires substantial expertise and experimental infrastructure to realize practical impact, responsible development and deployment are important considerations.

\bibliography{references}

@article{Lyu2019,
  title = {Ultra-large library docking for discovering new chemotypes},
  volume = {566},
  url = {http://dx.doi.org/10.1038/s41586-019-0917-9},
  DOI = {10.1038/s41586-019-0917-9},
  number = {7743},
  journal = {Nature},
  publisher = {Springer Science and Business Media LLC},
  author = {Lyu,  Jiankun and Wang,  Sheng and Balius,  Trent E. and Singh,  Isha and Levit,  Anat and Moroz,  Yurii S. and O’Meara,  Matthew J. and Che,  Tao and Algaa,  Enkhjargal and Tolmachova,  Kateryna and Tolmachev,  Andrey A. and Shoichet,  Brian K. and Roth,  Bryan L. and Irwin,  John J.},
  year = {2019},
  pages = {224–229}
}

@article{Graff2021,
  title = {Accelerating high-throughput virtual screening through molecular pool-based active learning},
  volume = {12},
  url = {http://dx.doi.org/10.1039/D0SC06805E},
  DOI = {10.1039/d0sc06805e},
  number = {22},
  journal = {Chemical Science},
  publisher = {Royal Society of Chemistry (RSC)},
  author = {Graff,  David E. and Shakhnovich,  Eugene I. and Coley,  Connor W.},
  year = {2021},
  pages = {7866–7881}
}

@inproceedings{kristiadi2024sober,
author = {Kristiadi, Agustinus and Strieth-Kalthoff, Felix and Skreta, Marta and Poupart, Pascal and Aspuru-Guzik, Al\'{a}n and Pleiss, Geoff},
title = {A sober look at LLMs for material discovery: are they actually good for bayesian optimization over molecules?},
url = {http://dx.doi.org/10.48550/arXiv.2402.05015},
DOI = {10.48550/arXiv.2402.05015},
year = {2024},
publisher = {JMLR.org},
booktitle = {Proceedings of the 41st International Conference on Machine Learning},
articleno = {1025},
numpages = {20},
location = {Vienna, Austria},
series = {ICML'24}
}

@article{Gorgulla2023,
  title = {AI-Enhanced Adaptive Virtual Screening Platform Enabling Exploration of 69 Billion Molecules Discovers Structurally Validated FSP1 Inhibitors},
  url = {http://dx.doi.org/10.1101/2023.04.25.537981},
  DOI = {10.1101/2023.04.25.537981},
  journal = {bioRxiv Preprint},
  author = {Gorgulla,  Christoph and Cecchini,  Domiziana and Nigam,  AkshatKumar and Tang,  Ming and Reis,  Joana and Koop,  Matt and Gottinger,  Andrea and Nicoll,  Callum Robert and Jayaraj,  Abhilash and Cinaroglu,  Suleyman Selim and Torner,  Ricarda and Seo,  Hyuk-Soo and Dhe-Paganon,  Sirano and Secker,  Christopher and Haddadnia,  Mohammad and Malets,  Yehor and Hasson,  Alexander and Das,  Krishna and Churion,  Kelly and Kim,  Jongwan and Li,  Minkai and Kumar,  Abhishek and Levin-Konigsberg,  Roni and Choi,  Eun-Bee and Shapiro,  Geoffrey and Cox,  Huel and Sebastian,  Luke and Braithwaite,  Chelsea and Bashyal,  Puspalata and Radchenko,  Dmytro S. and Kumar,  Aditya and Gehev,  Minko and Aquilanti,  Pierre-Yves and Gabb,  Henry and Alhossary,  Amr and Wagner,  Gerhard and Aspuru-Guzik,  Alan and Moroz,  Yurii S. and Kalodimos,  Charalampos G. and Fackeldey,  Konstantin and Mattevi,  Andrea and Arthanari,  Haribabu},
  year = {2023},
}

@misc{rdkit,
  doi = {10.5281/ZENODO.8053810},
  url = {https://zenodo.org/record/8053810},
  author = {Landrum,  Greg and Tosco,  Paolo and Kelley,  Brian and Rodriguez-Schmidt, Ricardo and Cosgrove,  David and Riniker, Sereina and Gedeck, Peter and Vianello,  Riccardo and Schneider, Nadine and Kawashima,  Eisuke and N,  Dan and Jones,  Gareth and Dalke,  Andrew and Cole,  Brian and Swain,  Matt and Turk,  Samo and Savelyev, Alexander and Vaucher,  Alain and Wójcikowski,  Maciej and Take, Ichiru and Probst,  Daniel and Ujihara,  Kazuya and Scalfani,  Vincent F. and Godin,  Guillaume and Lehtivarjo,  Juuso and Pahl,  Axel and Walker,  Rachel and Berenger, Francois and Biggs, Jason D.},
  title = {RDKit (Release 2023-03-2)},
  publisher = {Zenodo},
  year = {2023},
}

@article{OBoyle2011,
  title = {Open Babel: An open chemical toolbox},
  volume = {3},
  url = {http://dx.doi.org/10.1186/1758-2946-3-33},
  DOI = {10.1186/1758-2946-3-33},
  number = {1},
  journal = {Journal of Cheminformatics},
  publisher = {Springer Science and Business Media LLC},
  author = {O’Boyle,  Noel M and Banck,  Michael and James,  Craig A and Morley,  Chris and Vandermeersch,  Tim and Hutchison,  Geoffrey R},
  year = {2011},
}

@article{Yu2023,
  title = {Uni-Dock: GPU-Accelerated Docking Enables Ultralarge Virtual Screening},
  volume = {19},
  url = {http://dx.doi.org/10.1021/acs.jctc.2c01145},
  DOI = {10.1021/acs.jctc.2c01145},
  number = {11},
  journal = {Journal of Chemical Theory and Computation},
  publisher = {American Chemical Society (ACS)},
  author = {Yu,  Yuejiang and Cai,  Chun and Wang,  Jiayue and Bo,  Zonghua and Zhu,  Zhengdan and Zheng,  Hang},
  year = {2023},
  pages = {3336–3345}
}

@inproceedings{christofidellis2023unifying,
  title = 	 {Unifying Molecular and Textual Representations via Multi-task Language Modelling},
  url = {http://dx.doi.org/10.48550/arXiv.2301.12586},
  DOI = {10.48550/arXiv.2301.12586},
  author =       {Christofidellis, Dimitrios and Giannone, Giorgio and Born, Jannis and Winther, Ole and Laino, Teodoro and Manica, Matteo},
  booktitle = 	 {Proceedings of the 40th International Conference on Machine Learning (ICML 2023)},
  pages = 	 {6140--6157},
  year = 	 {2023},
  publisher =    {PMLR},
  url = 	 {https://proceedings.mlr.press/v202/christofidellis23a.html},
}

@article{2020t5,
  author  = {Colin Raffel and Noam Shazeer and Adam Roberts and Katherine Lee and Sharan Narang and Michael Matena and Yanqi Zhou and Wei Li and Peter J. Liu},
  title   = {Exploring the Limits of Transfer Learning with a Unified Text-to-Text Transformer},
  journal = {Journal of Machine Learning Research},
  year    = {2020},
  volume  = {21},
  number  = {140},
  pages   = {1-67},
  url     = {http://jmlr.org/papers/v21/20-074.html},
  DOI = {10.48550/arXiv.1910.10683}
}

@article{Auer2002,
  title = {Finite-time Analysis of the Multiarmed Bandit Problem},
  volume = {47},
  url = {http://dx.doi.org/10.1023/A:1013689704352},
  DOI = {10.1023/a:1013689704352},
  number = {2–3},
  journal = {Machine Learning},
  publisher = {Springer Science and Business Media LLC},
  author = {Auer,  Peter and Cesa-Bianchi,  Nicolò and Fischer,  Paul},
  year = {2002},
  pages = {235–256}
}

@book{Lattimore2020,
  title = {Bandit Algorithms},
  url = {http://dx.doi.org/10.1017/9781108571401},
  DOI = {10.1017/9781108571401},
  publisher = {Cambridge University Press},
  author = {Lattimore,  Tor and Szepesvári,  Csaba},
  year = {2020},
}

@inproceedings{laplace2021,
author = {Daxberger, Erik and Kristiadi, Agustinus and Immer, Alexander and Eschenhagen, Runa and Bauer, Matthias and Hennig, Philipp},
title = {Laplace redux – effortless Bayesian deep learning},
url = {http://dx.doi.org/10.48550/arXiv.2106.14806},
DOI = {10.48550/arXiv.2106.14806},
year = {2021},
publisher = {Curran Associates Inc.},
address = {Red Hook, NY, USA},
booktitle = {Proceedings of the 35th International Conference on Neural Information Processing Systems (NeurIPS 2021)},
pages = {20089 - 20103}
}

@InProceedings{Gal2016Dropout,
  title = 	 {Dropout as a Bayesian Approximation: Representing Model Uncertainty in Deep Learning},
  author = 	 {Gal, Yarin and Ghahramani, Zoubin},
  booktitle = 	 {Proceedings of the 33rd International Conference on Machine Learning (ICML 2016)},
  pages = 	 {1050--1059},
  year = 	 {2016},
  editor = 	 {Balcan, Maria Florina and Weinberger, Kilian Q.},
  volume = 	 {48},
  address = 	 {New York, New York, USA},
  publisher =    {PMLR},
  url = 	 {https://proceedings.mlr.press/v48/gal16.html},
  DOI = {10.48550/arXiv.1506.02142}
}

@inproceedings{Maddox2019SWAG,
  author = {Maddox, Wesley J. and Izmailov, Pavel and Garipov, Timur and Vetrov, Dmitry P. and Wilson, Andrew Gordon},
  title = {A Simple Baseline for {B}ayesian Uncertainty in Deep Learning},
  booktitle = {Proceedings of the 32nd International Conference on Neural Information Processing Systems (NeurIPS 2019)},
  pages = {13132--13143},
  year = {2019},
  url = {http://dx.doi.org/10.48550/arXiv.1902.02476},
  doi = {10.48550/arXiv.1902.02476}
}

@inproceedings{AdamOpt,
  author       = {Diederik P. Kingma and
                  Jimmy Ba},
  title        = {Adam: {A} Method for Stochastic Optimization},
  booktitle    = {3rd International Conference on Learning Representations (ICLR 2015)},
  year         = {2015},
  url          = {http://dx.doi.org/10.48550/arXiv.1412.6980},
  doi = {10.48550/arXiv.1412.6980}
}

@inproceedings{CosineAnnealingLR,
  author       = {Ilya Loshchilov and
                  Frank Hutter},
  title        = {{SGDR:} Stochastic Gradient Descent with Warm Restarts},
  booktitle    = {5th International Conference on Learning Representations (ICLR 2017)},
  year         = {2017},
  bibsource    = {dblp computer science bibliography, https://dblp.org},
  doi = {10.48550/arXiv.1608.03983},
  url = {http://dx.doi.org/10.48550/arXiv.1608.03983},
}

@inproceedings{
wang-henderson2023graph,
title={Graph Neural Network Powered Bayesian Optimization for Large Molecular Spaces},
author={Miles Wang-Henderson and Bartu Soyuer and Parnian Kassraie and Andreas Krause and Ilija Bogunovic},
booktitle={ICML 2023 Workshop on Structured Probabilistic Inference {\&} Generative Modeling},
year={2023},
url={https://openreview.net/forum?id=QIrgM7uybw}
}

@article{Restrepo2022,
  title = {Chemical space: limits,  evolution and modelling of an object bigger than our universal library},
  volume = {1},
  url = {http://dx.doi.org/10.1039/D2DD00030J},
  DOI = {10.1039/d2dd00030j},
  number = {5},
  journal = {Digital Discovery},
  publisher = {Royal Society of Chemistry (RSC)},
  author = {Restrepo,  Guillermo},
  year = {2022},
  pages = {568–585}
}

@article{Papidocha2026,
  title = {The elephant in the lab: synthesizability in generative small-molecule design},
  volume = {51},
  url = {http://dx.doi.org/10.1016/j.coche.2025.101217},
  DOI = {10.1016/j.coche.2025.101217},
  journal = {Current Opinion in Chemical Engineering},
  publisher = {Elsevier BV},
  author = {Papidocha,  Sven M and Burger,  Andreas and Bernales,  Varinia and Aspuru-Guzik,  Alán},
  year = {2026},
  pages = {101217}
}

@article{Shoichet2004,
  title = {Virtual screening of chemical libraries},
  volume = {432},
  url = {http://dx.doi.org/10.1038/nature03197},
  DOI = {10.1038/nature03197},
  number = {7019},
  journal = {Nature},
  publisher = {Springer Science and Business Media LLC},
  author = {Shoichet,  Brian K.},
  year = {2004},
  pages = {862–865}
}

@article{Morgan2011,
  title = {The cost of drug development: A systematic review},
  volume = {100},
  url = {http://dx.doi.org/10.1016/j.healthpol.2010.12.002},
  DOI = {10.1016/j.healthpol.2010.12.002},
  number = {1},
  journal = {Health Policy},
  publisher = {Elsevier BV},
  author = {Morgan,  Steve and Grootendorst,  Paul and Lexchin,  Joel and Cunningham,  Colleen and Greyson,  Devon},
  year = {2011},
  pages = {4–17}
}

@article{Reker2015,
  title = {Active-learning strategies in computer-assisted drug discovery},
  volume = {20},
  url = {http://dx.doi.org/10.1016/j.drudis.2014.12.004},
  DOI = {10.1016/j.drudis.2014.12.004},
  number = {4},
  journal = {Drug Discovery Today},
  publisher = {Elsevier BV},
  author = {Reker,  Daniel and Schneider,  Gisbert},
  year = {2015},
  pages = {458–465}
}

@article{Reker2019,
  title = {Practical considerations for active machine learning in drug discovery},
  volume = {32–33},
  url = {http://dx.doi.org/10.1016/j.ddtec.2020.06.001},
  DOI = {10.1016/j.ddtec.2020.06.001},
  journal = {Drug Discovery Today: Technologies},
  publisher = {Elsevier BV},
  author = {Reker,  Daniel},
  year = {2019},
  pages = {73–79}
}

@misc{enamine,
	author = {{Enamine Ltd.}},
	title = {{E}namine {R}{E}{A}{L} {D}atabase},
	url = {https://enamine.net/compound-collections/real-compounds/real-database},
}

@article{Hoffmann2019,
  title = {The next level in chemical space navigation: going far beyond enumerable compound libraries},
  volume = {24},
  url = {http://dx.doi.org/10.1016/j.drudis.2019.02.013},
  DOI = {10.1016/j.drudis.2019.02.013},
  number = {5},
  journal = {Drug Discovery Today},
  publisher = {Elsevier BV},
  author = {Hoffmann,  Torsten and Gastreich,  Marcus},
  year = {2019},
  pages = {1148–1156}
}

@article{Warr2022,
  title = {Exploration of Ultralarge Compound Collections for Drug Discovery},
  volume = {62},
  url = {http://dx.doi.org/10.1021/acs.jcim.2c00224},
  DOI = {10.1021/acs.jcim.2c00224},
  number = {9},
  journal = {Journal of Chemical Information and Modeling},
  publisher = {American Chemical Society (ACS)},
  author = {Warr,  Wendy A. and Nicklaus,  Marc C. and Nicolaou,  Christos A. and Rarey,  Matthias},
  year = {2022},
  pages = {2021–2034}
}

@inproceedings{
    Grigg2025,
    title={Active Learning on Synthons for Molecular Design},
    author={Tom George Grigg and Mason Burlage and Oliver Brook Scott and Dominique Sydow and Liam Wilbraham},
    booktitle={ICLR 2025 Workshop on Generative and Experimental Perspectives for Biomolecular Design},
    year={2025},
    url={https://openreview.net/forum?id=bLUw471nio}
}

@article{Kozyrev2025,
  title = {Active Learning to Select the Most Suitable Reagents and One-Step Organic Chemistry Reactions for Prioritizing Target-Specific Hits from Ultralarge Chemical Spaces},
  volume = {65},
  url = {http://dx.doi.org/10.1021/acs.jcim.4c02097},
  DOI = {10.1021/acs.jcim.4c02097},
  number = {2},
  journal = {Journal of Chemical Information and Modeling},
  publisher = {American Chemical Society (ACS)},
  author = {Kozyrev,  Vladimir and Sindt,  Fran\c{c}ois and Rognan,  Didier},
  year = {2025},
  pages = {693–704}
}

@article{Klarich2024,
  title = {Thompson Sampling – An Efficient Method for Searching Ultralarge Synthesis on Demand Databases},
  volume = {64},
  url = {http://dx.doi.org/10.1021/acs.jcim.3c01790},
  DOI = {10.1021/acs.jcim.3c01790},
  number = {4},
  journal = {Journal of Chemical Information and Modeling},
  publisher = {American Chemical Society (ACS)},
  author = {Klarich,  Kathryn and Goldman,  Brian and Kramer,  Trevor and Riley,  Patrick and Walters,  W. Patrick},
  year = {2024},
  pages = {1158–1171}
}

@article{Zhao2025,
  title = {Enhanced Thompson sampling by roulette wheel selection for screening ultralarge combinatorial libraries},
  volume = {17},
  url = {http://dx.doi.org/10.1186/s13321-025-01105-1},
  DOI = {10.1186/s13321-025-01105-1},
  number = {1},
  journal = {Journal of Cheminformatics},
  publisher = {Springer Science and Business Media LLC},
  author = {Zhao,  Hongtao and Nittinger,  Eva and Yu,  Melissa A. and Gathiaka,  Symon and Walters,  W. Patrick and Tyrchan,  Christian},
  year = {2025},
}

@article{Nazarova2025,
  title = {V-synthes2 - the Next Generation Tool for Structure-based Virtual Screening of Giga-scale Chemical Spaces},
  url = {http://dx.doi.org/10.21203/rs.3.rs-7782723/v1},
  DOI = {10.21203/rs.3.rs-7782723/v1},
  publisher = {Springer Science and Business Media LLC},
  author = {Nazarova,  Antonina L. and Sadybekov,  Anastasiia V. and Sadybekov,  Arman A. and Protopopov,  Mykola and Radchenko,  Dmytro S. and Moroz,  Yurii S. and Tarkhanova,  Olga O. and Katritch,  Vsevolod},
  year = {2025},
  journal = {ResearchSquare Preprint}
}

@article{Gorgulla2020,
  title = {An open-source drug discovery platform enables ultra-large virtual screens},
  volume = {580},
  url = {http://dx.doi.org/10.1038/s41586-020-2117-z},
  DOI = {10.1038/s41586-020-2117-z},
  number = {7805},
  journal = {Nature},
  publisher = {Springer Science and Business Media LLC},
  author = {Gorgulla,  Christoph and Boeszoermenyi,  Andras and Wang,  Zi-Fu and Fischer,  Patrick D. and Coote,  Paul W. and Padmanabha Das,  Krishna M. and Malets,  Yehor S. and Radchenko,  Dmytro S. and Moroz,  Yurii S. and Scott,  David A. and Fackeldey,  Konstantin and Hoffmann,  Moritz and Iavniuk,  Iryna and Wagner,  Gerhard and Arthanari,  Haribabu},
  year = {2020},
  pages = {663–668}
}

@article{PyzerKnapp2018,
  title = {Bayesian optimization for accelerated drug discovery},
  volume = {62},
  url = {http://dx.doi.org/10.1147/JRD.2018.2881731},
  DOI = {10.1147/jrd.2018.2881731},
  number = {6},
  journal = {IBM Journal of Research and Development},
  publisher = {IBM},
  author = {Pyzer-Knapp,  E. O.},
  year = {2018},
  pages = {2:1--2:7}
}

@article{bo-tutorial,
  doi = {10.48550/arXiv.1807.02811},
  url = {https://arxiv.org/abs/1807.02811},
  journal = {arXiv Preprint},
  author = {Frazier,  Peter I.},
  title = {A Tutorial on Bayesian Optimization},
  year = {2018},
}

@book{garnett_bayesoptbook_2023,
  author    = {Garnett, Roman},
  title     = {{Bayesian Optimization}},
  year      = {2023},
  publisher = {Cambridge University Press},
  url = {https://bayesoptbook.com/},
}

@article{Lyu2023,
  title = {Modeling the expansion of virtual screening libraries},
  volume = {19},
  url = {http://dx.doi.org/10.1038/s41589-022-01234-w},
  DOI = {10.1038/s41589-022-01234-w},
  number = {6},
  journal = {Nature Chemical Biology},
  publisher = {Springer Science and Business Media LLC},
  author = {Lyu,  Jiankun and Irwin,  John J. and Shoichet,  Brian K.},
  year = {2023},
  pages = {712–718}
}

@article{Gloriam2019,
  title = {Bigger is better in virtual drug screens},
  volume = {566},
  url = {http://dx.doi.org/10.1038/d41586-019-00145-6},
  DOI = {10.1038/d41586-019-00145-6},
  number = {7743},
  journal = {Nature},
  publisher = {Springer Science and Business Media LLC},
  author = {Gloriam,  David E.},
  year = {2019},
  pages = {193–194}
}

@book{Settles2012,
  title = {Active Learning},
  url = {http://dx.doi.org/10.1007/978-3-031-01560-1},
  DOI = {10.1007/978-3-031-01560-1},
  journal = {Synthesis Lectures on Artificial Intelligence and Machine Learning},
  publisher = {Springer International Publishing},
  author = {Settles,  Burr},
  year = {2012}
}

@inproceedings{Garnett2012,
author = {Garnett, Roman and Krishnamurthy, Yamuna and Xiong, Xuehan and Schneider, Jeff and Mann, Richard},
title = {Bayesian optimal active search and surveying},
doi = {10.48550/arXiv.1206.6406},
url = {http://dx.doi.org/10.48550/arXiv.1206.6406},
year = {2012},
publisher = {Omnipress},
address = {Madison, WI, USA},
booktitle = {Proceedings of the 29th International Conference on Machine Learning (ICML 2012)},
pages = {843–850},
numpages = {8},
location = {Edinburgh, Scotland},
}

@article{Garnett2015,
  title = {Introducing the ‘active search’ method for iterative virtual screening},
  volume = {29},
  url = {http://dx.doi.org/10.1007/s10822-015-9832-9},
  DOI = {10.1007/s10822-015-9832-9},
  number = {4},
  journal = {Journal of Computer-Aided Molecular Design},
  publisher = {Springer Science and Business Media LLC},
  author = {Garnett,  Roman and G\"{a}rtner,  Thomas and Vogt,  Martin and Bajorath,  J\"{u}rgen},
  year = {2015},
  pages = {305–314}
}

@misc{Jiang2018,
  doi = {10.48550/ARXIV.1811.08871},
  url = {https://arxiv.org/abs/1811.08871},
  author = {Jiang,  Shali and Malkomes,  Gustavo and Moseley,  Benjamin and Garnett,  Roman},
  title = {Efficient nonmyopic active search with applications in drug and materials discovery},
  publisher = {arXiv},
  year = {2018},
  copyright = {arXiv.org perpetual,  non-exclusive license}
}

@article{Robbins1952,
  title = {Some aspects of the sequential design of experiments},
  volume = {58},
  url = {http://dx.doi.org/10.1090/S0002-9904-1952-09620-8},
  DOI = {10.1090/s0002-9904-1952-09620-8},
  number = {5},
  journal = {Bulletin of the American Mathematical Society},
  publisher = {American Mathematical Society (AMS)},
  author = {Robbins,  Herbert},
  year = {1952},
  pages = {527–535}
}

@article{Eisenhuth2025,
  title = {Ultra-large library screening with an evolutionary algorithm in Rosetta (REvoLd)},
  volume = {8},
  url = {http://dx.doi.org/10.1038/s42004-025-01758-x},
  DOI = {10.1038/s42004-025-01758-x},
  number = {1},
  journal = {Communications Chemistry},
  publisher = {Springer Science and Business Media LLC},
  author = {Eisenhuth,  Paul and Liessmann,  Fabian and Moretti,  Rocco and Meiler,  Jens},
  year = {2025},
}

@article{Sadybekov2021,
  title = {Synthon-based ligand discovery in virtual libraries of over 11 billion compounds},
  volume = {601},
  url = {http://dx.doi.org/10.1038/s41586-021-04220-9},
  DOI = {10.1038/s41586-021-04220-9},
  number = {7893},
  journal = {Nature},
  publisher = {Springer Science and Business Media LLC},
  author = {Sadybekov,  Arman A. and Sadybekov,  Anastasiia V. and Liu,  Yongfeng and Iliopoulos-Tsoutsouvas,  Christos and Huang,  Xi-Ping and Pickett,  Julie and Houser,  Blake and Patel,  Nilkanth and Tran,  Ngan K. and Tong,  Fei and Zvonok,  Nikolai and Jain,  Manish K. and Savych,  Olena and Radchenko,  Dmytro S. and Nikas,  Spyros P. and Petasis,  Nicos A. and Moroz,  Yurii S. and Roth,  Bryan L. and Makriyannis,  Alexandros and Katritch,  Vsevolod},
  year = {2021},
  pages = {452–459}
}

@article{Grygorenko2020,
  title = {Generating Multibillion Chemical Space of Readily Accessible Screening Compounds},
  volume = {23},
  url = {http://dx.doi.org/10.1016/j.isci.2020.101681},
  DOI = {10.1016/j.isci.2020.101681},
  number = {11},
  journal = {iScience},
  publisher = {Elsevier BV},
  author = {Grygorenko,  Oleksandr O. and Radchenko,  Dmytro S. and Dziuba,  Igor and Chuprina,  Alexander and Gubina,  Kateryna E. and Moroz,  Yurii S.},
  year = {2020},
  pages = {101681}
}

@article{Bong2008,
  title = {Structural studies of human brain‐type creatine kinase complexed with the ADP–Mg2+NO3-–creatine transition‐state analogue complex},
  volume = {582},
  url = {http://dx.doi.org/10.1016/j.febslet.2008.10.039},
  DOI = {10.1016/j.febslet.2008.10.039},
  number = {28},
  journal = {FEBS Letters},
  publisher = {Wiley},
  author = {Bong,  Seoung Min and Moon,  Jin Ho and Nam,  Ki Hyun and Lee,  Ki Seog and Chi,  Young Min and Hwang,  Kwang Yeon},
  year = {2008},
  pages = {3959–3965}
}

@article{Maspero2025,
  title = {Structure-based design of potent and selective inhibitors of the {HECT} ligase {NEDD4}},
  volume = {8},
  url = {http://dx.doi.org/10.1038/s42004-025-01557-4},
  DOI = {10.1038/s42004-025-01557-4},
  number = {1},
  journal = {Communications Chemistry},
  publisher = {Springer Science and Business Media LLC},
  author = {Maspero,  Elena and Cappa,  Anna and Weber,  Janine and Trifirò,  Paolo and Amici,  Raffaella and Bruno,  Agostino and Fagà,  Giovanni and Cecatiello,  Valentina and Fattori,  Raimondo and Leuzzi,  Brian and Taibi,  Vincenzo and Meroni,  Giuseppe and Pasi,  Maurizio and Romussi,  Alessia and Sartori,  Luca and Villa,  Manuela and Vultaggio,  Stefania and Cirò,  Marco and Soffientini,  Paolo and Lombardo,  Lierin and Dahe,  Shakti and Bachi,  Angela and Varasi,  Mario and Rossi,  Mario and Pasqualato,  Sebastiano and Mercurio,  Ciro and Polo,  Simona},
  year = {2025},
}

@article{Liu2025,
  title = {The impact of library size and scale of testing on virtual screening},
  volume = {21},
  url = {http://dx.doi.org/10.1038/s41589-024-01797-w},
  DOI = {10.1038/s41589-024-01797-w},
  number = {7},
  journal = {Nature Chemical Biology},
  publisher = {Springer Science and Business Media LLC},
  author = {Liu,  Fangyu and Mailhot,  Olivier and Glenn,  Isabella S. and Vigneron,  Seth F. and Bassim,  Violla and Xu,  Xinyu and Fonseca-Valencia,  Karla and Smith,  Matthew S. and Radchenko,  Dmytro S. and Fraser,  James S. and Moroz,  Yurii S. and Irwin,  John J. and Shoichet,  Brian K.},
  year = {2025},
  pages = {1039–1045}
}

@inproceedings{Levine2017,
author = {Levine, Nir and Crammer, Koby and Mannor, Shie},
title = {Rotting Bandits},
doi = {10.48550/arXiv.1702.07274},
url = {http://dx.doi.org/10.48550/arXiv.1702.07274},
year = {2017},
publisher = {Curran Associates Inc.},
address = {Red Hook, NY, USA},
booktitle = {Proceedings of the 31st International Conference on Neural Information Processing Systems (NeurIPS 2017)},
pages = {3077–3086},
}

@article{Naik2015,
    author = {Naik, Maruti and Raichurkar, Anandkumar and Bandodkar, Balachandra S. and Varun, Begur V. and Bhat, Shantika and Kalkhambkar, Rajesh and Murugan, Kannan and Menon, Rani and Bhat, Jyothi and Paul, Beena and Iyer, Harini and Hussein, Syeed and Tucker, Julie A. and Vogtherr, Martin and Embrey, Kevin J. and McMiken, Helen and Prasad, Swati and Gill, Adrian and Ugarkar, Bheemarao G. and Venkatraman, Janani and Read, Jon and Panda, Manoranjan},
    title = {Structure Guided Lead Generation for M. tuberculosis Thymidylate Kinase (Mtb TMK): Discovery of 3-Cyanopyridone and 1,6-Naphthyridin-2-one as Potent Inhibitors},
    journal = {Journal of Medicinal Chemistry},
    volume = {58},
    number = {2},
    pages = {753-766},
    year = {2015},
    DOI = {10.1021/jm5012947},
    url = {https://doi.org/10.1021/jm5012947}
}
\bibliographystyle{icml2026}

\newpage
\appendix
\onecolumn
\raggedbottom

\section{Additional Implementation Details}

\subsection{Fair Budgeting}

Importantly, although clustering reduces the average search space per round from $N$ to approximately $N/K$, we do \emph{not} reduce the batch size when using \textsc{BoBa}. 
One might consider evaluating only $B/K$ molecules per iteration within each subspace and increasing the number of rounds accordingly; however, this would equalize the total number of surrogate inferences between \textsc{BoBa} and standard BO, obscuring the regime we aim to study.

Instead, we fix both $T$ and $B$ across all methods. 
As a result, \textsc{BoBa} performs approximately $K$-fold fewer surrogate inferences than standard BO, directly exposing the tradeoff between inference cost and optimization performance.

\subsection{Runtime Breakdown}\label{appendix:runtime}

To complement the inference-count analysis in the main text, Table~\ref{table:runtime} reports wall-clock measurements for preprocessing and optimization on the ZINC--AmpC scaling benchmark.
The measurements separate one-time library preprocessing costs, namely embedding computation and clustering, from the repeated optimization loop.
Embedding computation dominates total runtime, but this cost is amortized across downstream targets because the same molecular representations and partitions can be reused.
Within a fixed optimization campaign, the dominant avoidable cost is surrogate inference, which is precisely the component reduced by restricting acquisition evaluation to the selected partition.

\begin{table}[h]
\centering
\small
\begin{tabular}{lrrrrrrrrrr}
\toprule
Dataset & Emb. & $K=5$ & $K=10$ & $K=25$ & $K=100$ & BO & BoBa-5 & BoBa-10 & BoBa-25 & BoBa-100 \\
\midrule
96M & 12774 & 541 & 649 & 810 & 855 & 2720 & 2137 & 2258 & 2318 & 2267 \\
10M & 1281 & 88 & 89 & 96 & 99 & 469 & 361 & 286 & 321 & 319 \\
1M & 129 & 3 & 12 & 14 & 18 & 57 & 50 & 41 & 37 & 39 \\
\bottomrule
\end{tabular}
\caption{\textbf{Wall-clock runtime breakdown for the ZINC--AmpC scaling benchmark.}
All values are reported in seconds. ``Emb.'' denotes embedding computation, the columns labeled by $K$ denote clustering/tranching time for the corresponding number of partitions, and the final columns report optimization-loop runtime for full-library BO and \textsc{BoBa}. 
Model training and inference were performed on an NVIDIA GH200 Grace Hopper chip.}
\label{table:runtime}
\end{table}

\subsection{Multi-Armed Bandit Algorithms} \label{app:mab}

In this work, each clustered subspace $\mathcal{X}_k$ is treated as an arm in a multi-armed bandit (MAB) problem. 
At iteration $t$, selecting arm $k$ yields a scalar reward $r_{k,t}$, defined as the average docking score of the batch evaluated from that subspace.
We maintain, for each arm $k$, the number of times it has been selected $n_k(t)$ and its empirical mean reward
\[
\hat{\mu}_k(t) = \frac{1}{n_k(t)} \sum_{i=1}^{n_k(t)} r_{k,i}.
\]

\paragraph{Initialization.}
All algorithms use a single-round initialization strategy: the initial batch is distributed across arms, with $B/K$ molecules seeded from each partition when $B$ is divisible by $K$.
This seeding strategy initializes every arm without spending one full optimization round per arm, which is important when $T$ is small relative to $K$.

We consider the following bandit strategies.

\subsubsection{UCB1}

Upper Confidence Bound (UCB1) selects the arm that maximizes an optimism-adjusted estimate of the mean reward \citep{Auer2002}:
\[
k_t = \arg\max_{k \in \{1,\dots,K\}} \left[ \hat{\mu}_k(t-1) + c \sqrt{\frac{2 \log t}{n_k(t-1)}} \right],
\]
where $c > 0$ is a tunable exploration constant (set to $c=1$ in all experiments unless otherwise stated).
Arms are initialized by the single-round seeding strategy described above before the UCB1 rule is applied.

\subsubsection{$\epsilon$-Greedy}

The $\epsilon$-greedy strategy selects arms according to
\[
k_t =
\begin{cases}
\arg\max_k \hat{\mu}_k(t-1), & \text{with probability } 1 - \epsilon, \\
\text{Uniform}(\{1,\dots,K\}), & \text{with probability } \epsilon,
\end{cases}
\]
where $\epsilon \in [0,1]$ controls the exploration rate. 
We report results for multiple values of $\epsilon$.

\subsubsection{Softmax Sampling}

Softmax (Boltzmann) exploration samples arms stochastically according to their empirical mean rewards:
\[
P(k_t = k) = \frac{\exp(\tau \, \hat{\mu}_k(t-1))}{\sum_{j=1}^K \exp(\tau \, \hat{\mu}_j(t-1))}.
\]
Here $\tau > 0$ is an inverse-temperature parameter controlling the sharpness of the distribution.
Larger $\tau$ increasingly concentrates probability mass on the empirically best arm, while smaller $\tau$ approaches uniform sampling.
Note that we parameterize softmax using $\exp(\tau x)$ rather than the more common $\exp(x/\tau)$; under this convention, larger $\tau$ corresponds to lower stochasticity.

\subsection{Bayesian Optimization} \label{app:bo_details}

All screening methods in this work, including standard Bayesian optimization and \textsc{BoBa}, are built on a common surrogate modeling and acquisition framework.
This shared framework ensures that observed differences arise from the allocation strategy rather than from differences in modeling capacity.

\subsubsection{Surrogate Model}

We model the unknown objective function $f(x)$ using Bayesian neural networks (BNNs) trained on the growing dataset $\mathcal{D}_t = \{(x_i, y_i)\}_{i=1}^t$.
Rather than performing fully Bayesian training, we adopt a post-hoc Laplace approximation to the posterior over network weights \citep{laplace2021}, which has recently been shown to provide accurate and computationally efficient uncertainty estimates for deep models.

Concretely, we first train a deterministic neural network by minimizing mean squared error on $\mathcal{D}_t$, which learns to predict docking scores from some given representation space $\phi(x) \in \mathbb{R}^d$.
Let $\hat{\theta}_t$ denote the resulting parameters.
We then approximate the posterior $p(\theta \mid \mathcal{D}_t)$ by a Gaussian centered at $\hat{\theta}_t$ with covariance given by the inverse Hessian of the negative log-likelihood,
\[
p(\theta \mid \mathcal{D}_t) \approx \mathcal{N}(\hat{\theta}_t, H_t^{-1}),
\]
where $H_t$ is estimated using a block-diagonal or Kronecker-factored approximation.
This approximation yields a predictive distribution for each candidate molecule,
\[
p(f(x) \mid \mathcal{D}_t) \approx \mathcal{N}(\mu_t(x), \sigma_t^2(x)),
\]
which provides both predictive means and calibrated epistemic uncertainties.

Unless otherwise stated, the same surrogate architecture, training protocol, and Laplace approximation procedure are used across all experiments.

\subsubsection{Acquisition Function}

Candidate selection is driven by the Upper Confidence Bound (UCB) acquisition function,
\[
a_t(x) = \mu_t(x) + \beta_t \, \sigma_t(x),
\]
where $\mu_t(x)$ and $\sigma_t(x)$ denote the predictive mean and standard deviation of the surrogate model, and $\beta_t > 0$ controls the exploration--exploitation tradeoff.

In standard Bayesian optimization, $a_t(x)$ is evaluated over the entire library.
In \textsc{BoBa}, the same acquisition function is used, but inference is restricted to the subspace selected by the bandit layer.
Thus, all methods share identical local decision rules; only the scope of inference and the global allocation of computation differ.

\subsubsection{Training and Update Protocol}

The surrogate model is retrained (or warm-started and updated) at each iteration using all available observations in $\mathcal{D}_t$.

\paragraph{Model architecture.}
The surrogate model is a fully connected feedforward neural network with two hidden layers of sizes 512 and 128, respectively, and ReLU nonlinearities.
The network maps molecular feature vectors to a scalar prediction of docking score.

\paragraph{Optimization.}
All networks are trained using the Adam optimizer \citep{AdamOpt} with learning rate $1\times10^{-3}$ and weight decay $1\times10^{-4}$.
We employ a cosine annealing learning rate schedule \citep{CosineAnnealingLR} over the course of training.
Each surrogate update consists of 500 training epochs with a batch size of 8192.

\paragraph{Laplace approximation.}
After deterministic training, we construct a post-hoc Bayesian neural network using a Laplace approximation over the final layer weights.
We use a last-layer Laplace approximation \citep{laplace2021} and refine the posterior for an additional 100 post-hoc optimization epochs.
Predictive means and uncertainties are obtained from this approximate posterior and used to compute UCB acquisition values.

\subsection{Uncertainty Estimation Ablation}\label{appendix:uncertainty}

\begin{figure*}[p]
    \centering
    \includegraphics[width=0.85\linewidth]{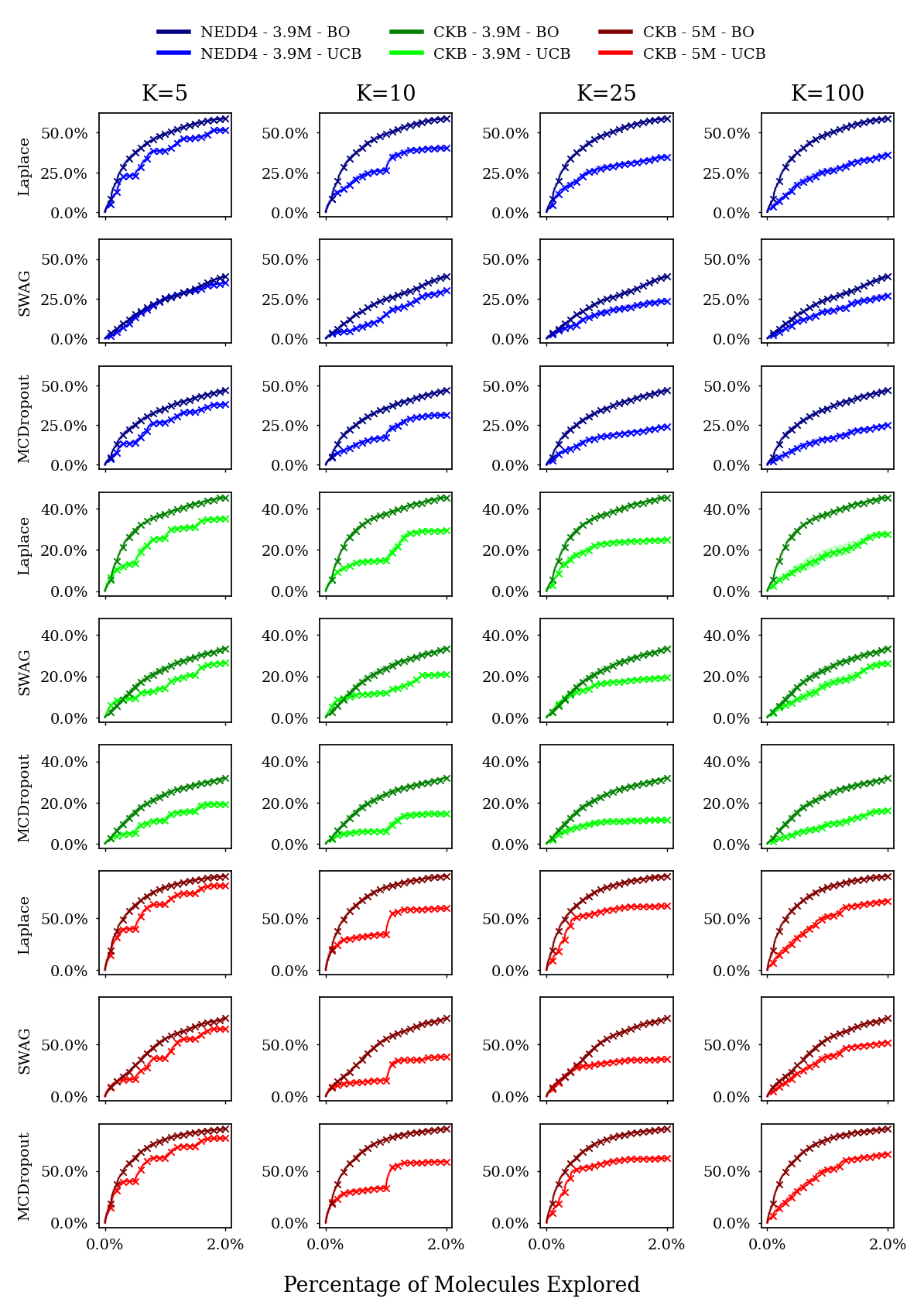}
    \caption{\textbf{Effect of uncertainty estimation on optimization performance.}
    Optimization trajectories for full-library BO and \textsc{BoBa} using Laplace approximation, SWAG, and MC dropout uncertainty estimates.
    Curves report retrieval of top-ranked candidates as a function of the number of molecules explored across \textsc{Enamine-S-3.9M} docked against NEDD4, \textsc{Enamine-S-3.9M} docked against CKB, and \textsc{Enamine-5M} docked against CKB.
    Columns correspond to different values of $K$.}
    \label{fig:uncertainty_ablation}
\end{figure*}

Because the acquisition function depends directly on predictive uncertainty, we evaluated whether the conclusions are specific to the Laplace approximation used in the main experiments.
Figure~\ref{fig:uncertainty_ablation} compares the Laplace approximation against SWAG \citep{Maddox2019SWAG} and MC dropout \citep{Gal2016Dropout} for both full-library BO and \textsc{BoBa} across multiple libraries, targets, and values of $K$.
Across these settings, the Laplace approximation provides the strongest or most consistent retrieval trajectories, particularly for full-library BO.
The same qualitative trend is retained in \textsc{BoBa}, although performance varies more strongly across libraries and cluster granularities.
These results support the use of the Laplace approximation as the default uncertainty estimator in the main benchmarks, while leaving more systematic uncertainty-calibration studies as a promising direction for future work.

\newpage

\section{Physiochemical Descriptors} \label{descriptors}

List of physiochemical properties used as heuristic features: aromatic proportion, aromatic ring count, atom count, bond count, chiral center count, doublebond stereoisomer count, electronegative atom count, formal charge, fraction of sp\textsuperscript{3} carbon atoms, halogen atom count, hydrogen bond acceptor count, hydrogen bond donor count, heavy atom count, logD, logP, logS, molecular refractivity, molecular weight, negative charge count, nitrogen and oxygen atom count, positive charge count, Quantitative Estimate of Druglikeness (QED), ring count, rotatable bond count, sulfur atom count, Topological Polar Surface Area (TPSA). For the 5 million compounds, the properties were obtained from \citep{Gorgulla2023}. The quantitative properties were selected from all properties calculated in \citep{Gorgulla2023}.

\newpage

\section{Additional Results}\label{appendix:results}

\begin{figure*}[h]
    \centering
    \begin{subfigure}{0.48\linewidth}
        \centering
        \includegraphics[width=\linewidth]{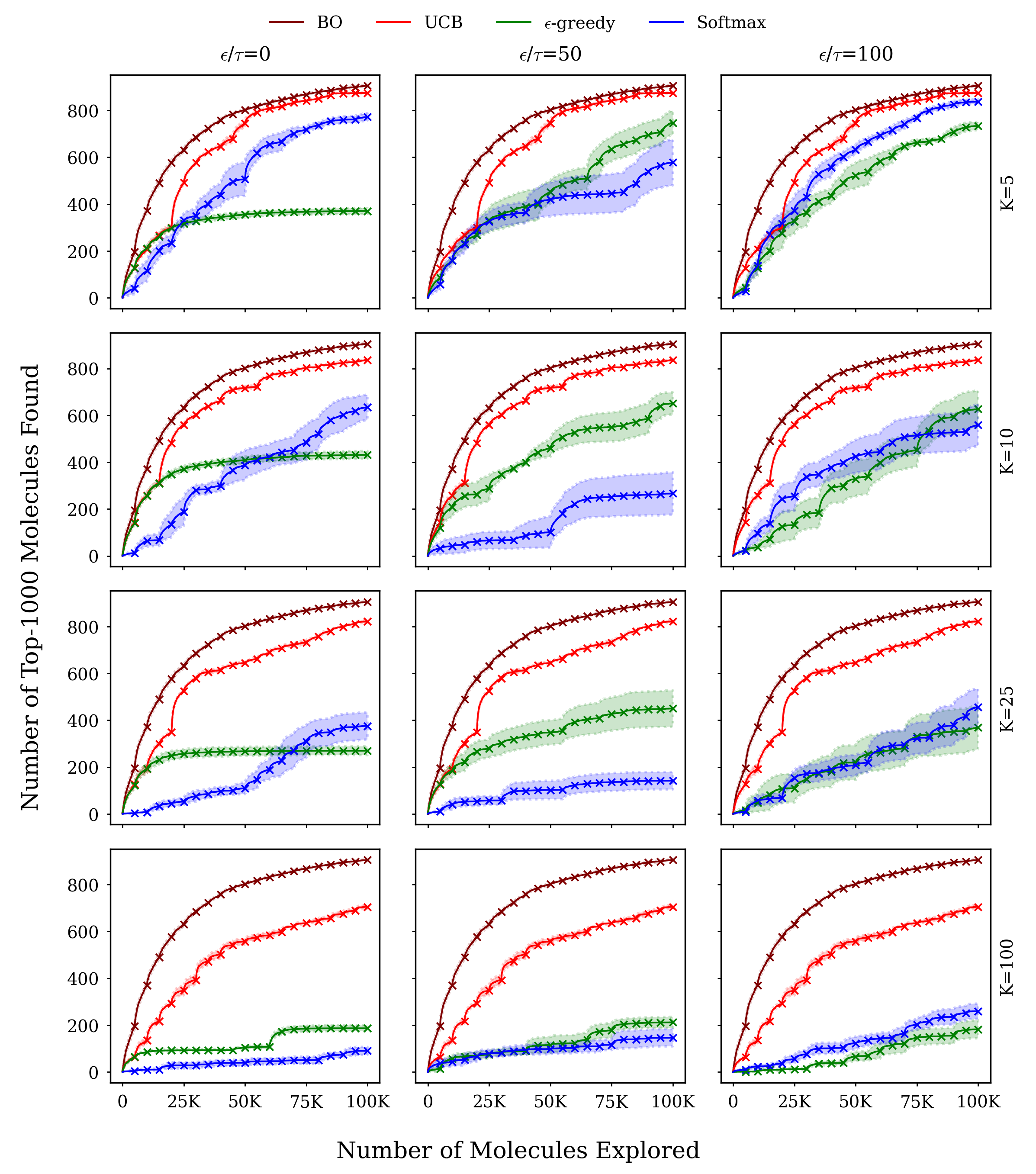}
        \label{fig:ckb5m-topK_a}
    \end{subfigure}
    \hfill
    \begin{subfigure}{0.48\linewidth}
        \centering
        \includegraphics[width=\linewidth]{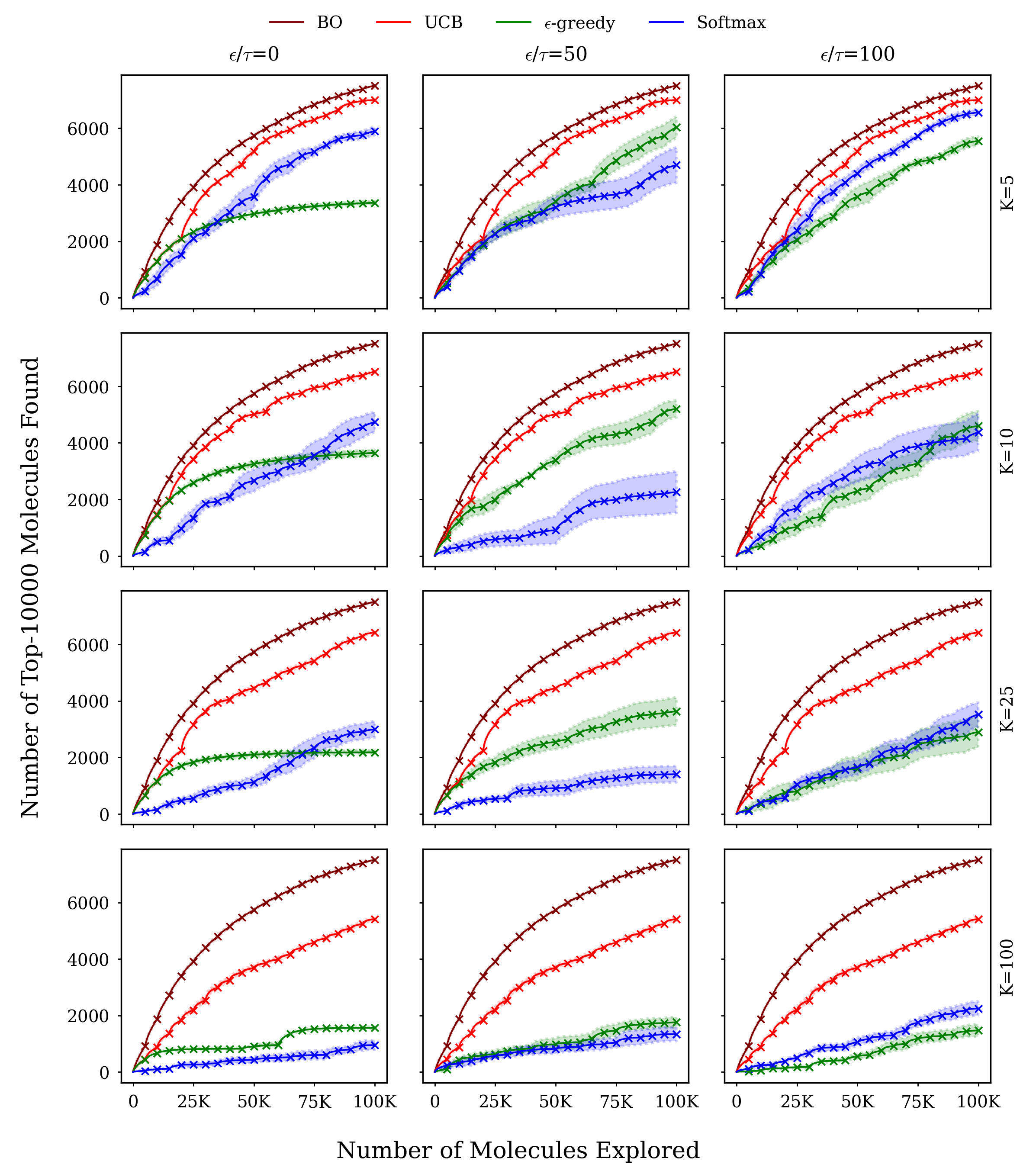}
        \label{fig:ckb5m-topK_b}
    \end{subfigure}
    \caption{Trajectories of top-1000 and top-10000 molecules recovered for the \textsc{Enamine-5M} library (CKB).}
    \label{fig:ckb5m-topK}
\end{figure*}

\begin{figure*}[h]
    \centering
    \begin{subfigure}{0.48\linewidth}
        \centering
        \includegraphics[width=\linewidth]{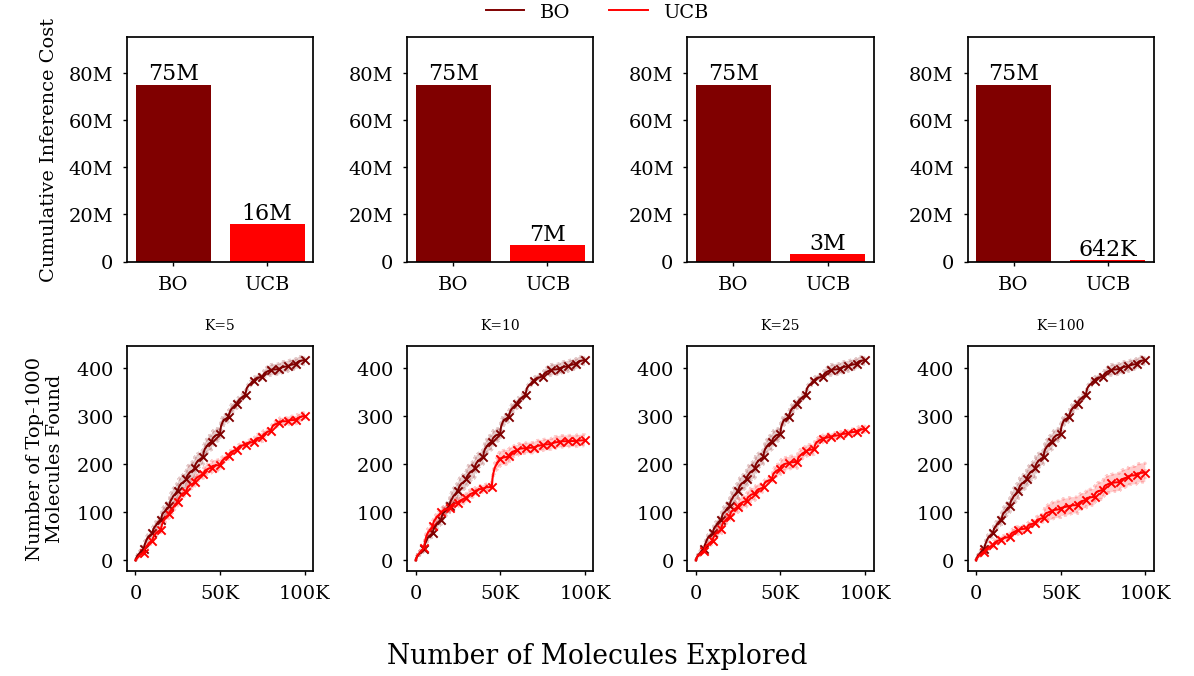}
        \label{fig:ckb3.9m-topK_a}
    \end{subfigure}
    \hfill
    \begin{subfigure}{0.48\linewidth}
        \centering
        \includegraphics[width=\linewidth]{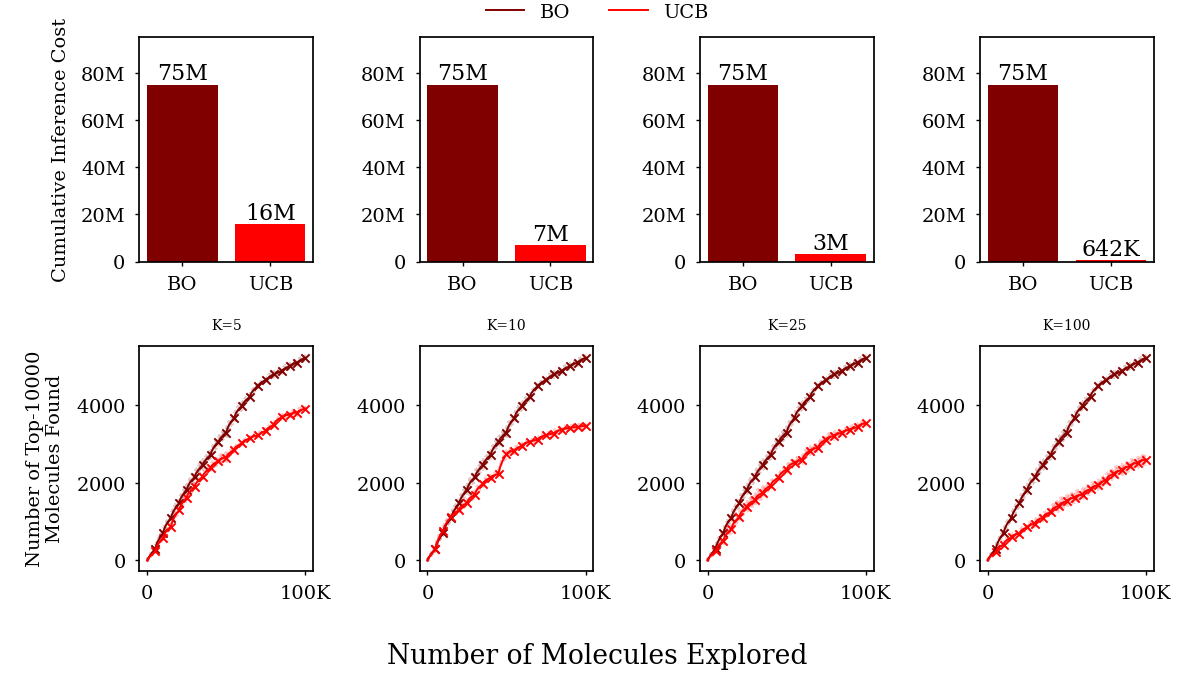}
        \label{fig:ckb3.9m-topK_b}
    \end{subfigure}
    \caption{Trajectories of top-1000 and top-10000 molecules recovered for the \textsc{Enamine-S-3.9M} (CKB) library}
    \label{fig:ckb3.9m-topK}
\end{figure*}

\begin{figure*}
    \centering
    \begin{subfigure}{0.48\linewidth}
        \centering
        \includegraphics[width=\linewidth]{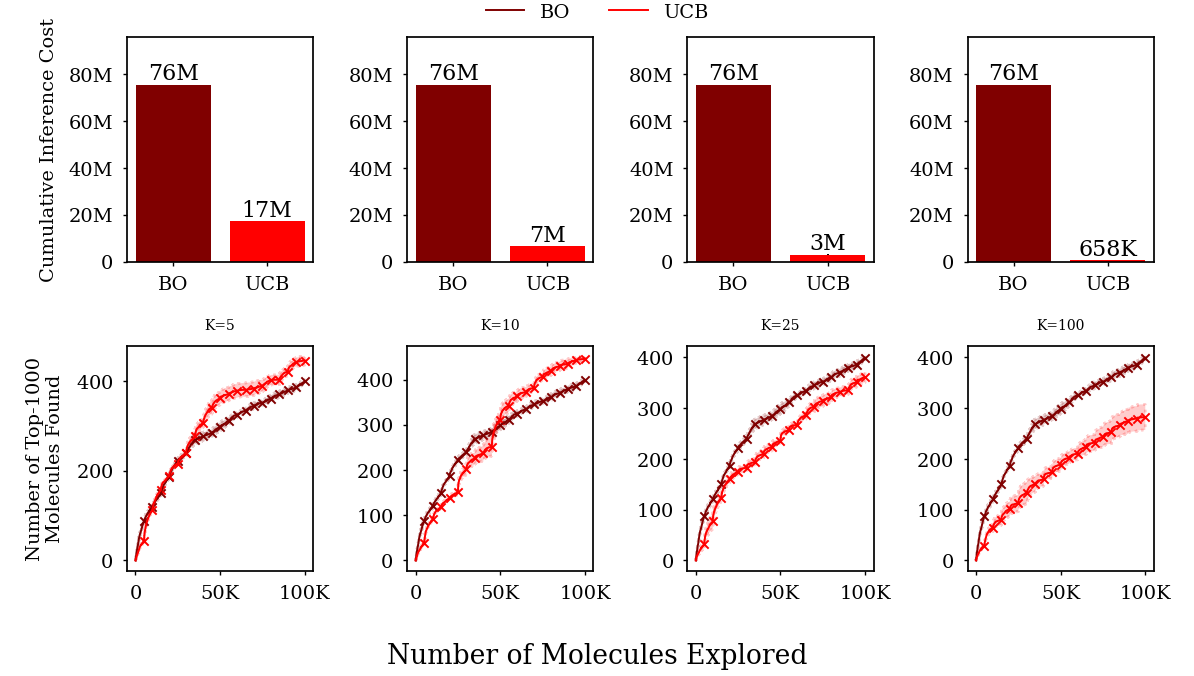}
        \label{fig:nedd3.9m-topK_a}
    \end{subfigure}
    \hfill
    \begin{subfigure}{0.48\linewidth}
        \centering
        \includegraphics[width=\linewidth]{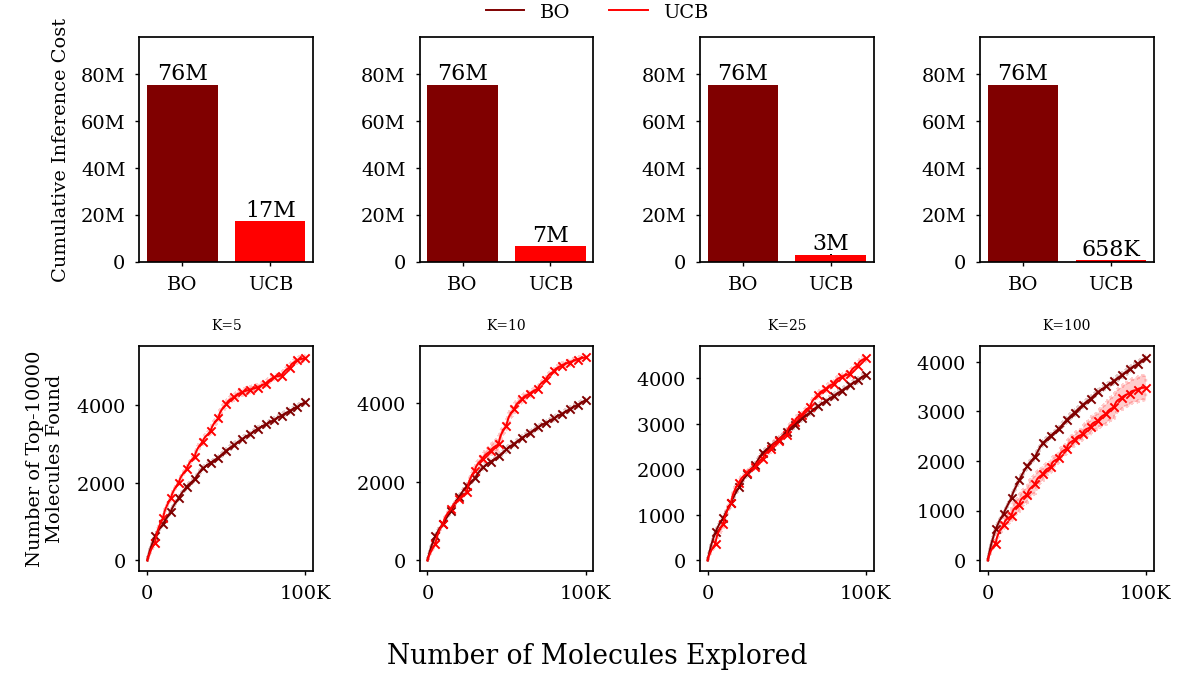}
        \label{fig:nedd3.9m-topK_b}
    \end{subfigure}
    \caption{Trajectories of top-1000 and top-10000 molecules recovered for the \textsc{Enamine-S-3.9M} (NEDD4) library}
    \label{fig:nedd3.9m-topK}
\end{figure*}

\newpage

\section{Arm Selection Frequencies}\label{appendix:arm-selection}

\begin{figure*}[h]
    \centering
    \includegraphics[width=0.65\linewidth]{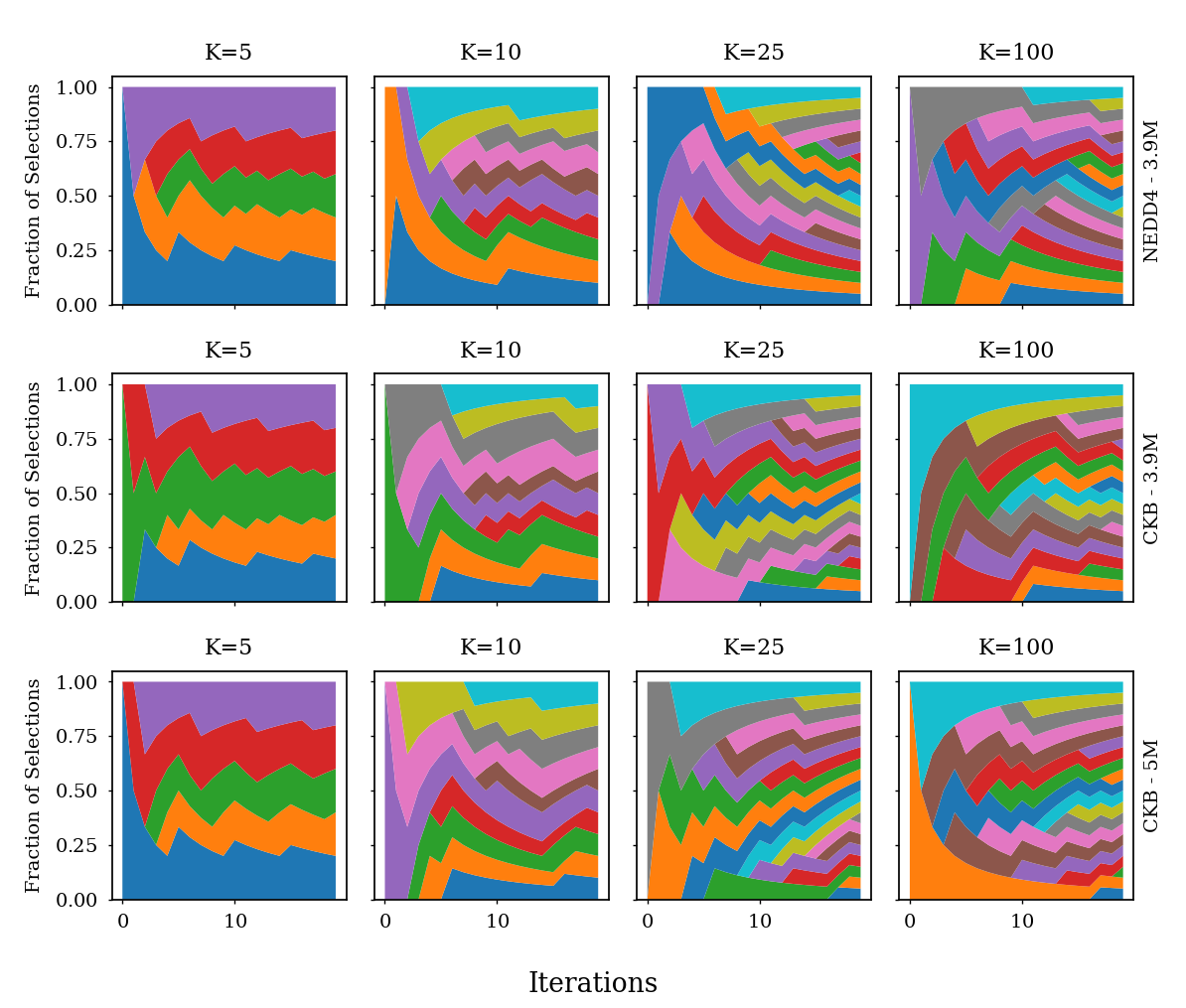}
    \caption{\textbf{Per-arm selection frequencies for \textsc{BoBa} with UCB1.}
    Each panel reports the fraction of selections assigned to each arm over optimization iterations, with arms color-coded by partition.
    Experiments are shown for \textsc{Enamine-S-3.9M} docked against NEDD4, \textsc{Enamine-S-3.9M} docked against CKB, and \textsc{Enamine-5M} docked against CKB, across $K \in \{5,10,25,100\}$.}
    \label{fig:arm-selection}
\end{figure*}

To assess whether UCB1 collapses onto a small number of initially high-reward partitions, we analyzed the fraction of selections assigned to each arm over the course of optimization.
Figure~\ref{fig:arm-selection} shows that UCB1 continues to distribute selections across multiple arms, even when the number of partitions is large.
This behavior is consistent with the optimism bonus in UCB1, which keeps under-sampled partitions competitive until their empirical rewards are sufficiently well characterized.
In the tested optimization budgets, we do not observe strong evidence of a winner's curse in which one early high-performing arm monopolizes the campaign after its best candidates are depleted.

\newpage

\section{Cost--Regret Tradeoff Analysis}\label{appendix:theory}

This section provides a simplified theoretical analysis of the central cost--performance tradeoff in \textsc{BoBa}.
The goal is not to fully model the molecular optimization problem, but to formalize the role of the number of clusters $K$ as a control parameter.

\paragraph{Setup.}
Consider a library of size $N$ partitioned into $K$ approximately equal clusters.
At each of $T$ optimization rounds, full-library BO performs surrogate inference over all $N$ candidates, whereas \textsc{BoBa} performs inference over approximately $N/K$ candidates after selecting one arm.
Thus, up to constants and batch-size factors shared across methods, the cumulative inference cost of \textsc{BoBa} scales as
\[
C_{\mathrm{inf}}(K) \asymp \frac{NT}{K}.
\]
Increasing $K$ therefore reduces inference cost, but it also makes the bandit allocation problem harder because the algorithm must identify promising regions among more arms.

\paragraph{Bandit term.}
Under a standard stochastic bandit abstraction, the regret of UCB-type allocation over $K$ arms admits gap-dependent logarithmic bounds and gap-free bounds of order
\[
R_{\mathrm{bandit}}(K,T) = \tilde{\mathcal{O}}(\sqrt{KT}),
\]
where logarithmic factors are suppressed.
This term captures the loss from allocating rounds to suboptimal partitions before the bandit has confidently identified high-yield regions.
The approximation ignores within-partition surrogate error and the mild non-stationarity induced by depletion of high-scoring candidates, but it isolates the effect of partition granularity.

\paragraph{Joint objective.}
Let $\lambda > 0$ denote the relative importance of inference cost compared with allocation regret.
A simplified objective is
\[
J(K) = \lambda \frac{NT}{K} + \sqrt{KT}.
\]
The first term decreases with $K$, while the second increases with $K$.

Treating $K$ as continuous, we can minimize $J$ with respect to $K$ by differentiation:
\[
\frac{dJ}{dK}
= -\lambda NTK^{-2} + \frac{1}{2}\sqrt{T}K^{-1/2}.
\]
Setting this derivative to zero yields
\[
K^{3/2} = 2\lambda N\sqrt{T},
\]
and therefore
\[
K^\ast \asymp N^{2/3}T^{1/3},
\]
up to constants and logarithmic factors.

\paragraph{Interpretation.}
This scaling law formalizes the empirical behavior observed in the main text.
Larger libraries justify finer partitions because the inference savings from reducing the queried fraction of the library grow with $N$.
At the same time, the optimal $K$ grows sublinearly, reflecting the increasing statistical burden of choosing among many arms.
In practice, the constant hidden in the scaling depends on hardware, surrogate architecture, batch size, cluster balance, target difficulty, and the quality of the molecular representation.
The analysis should therefore be interpreted as a design principle rather than a prescription for a single universal value of $K$.

\newpage

\section{Distribution of Docking Scores}\label{appendix:score-dist}

\begin{figure}[H]
    \centering
    \includegraphics[width=1\linewidth]{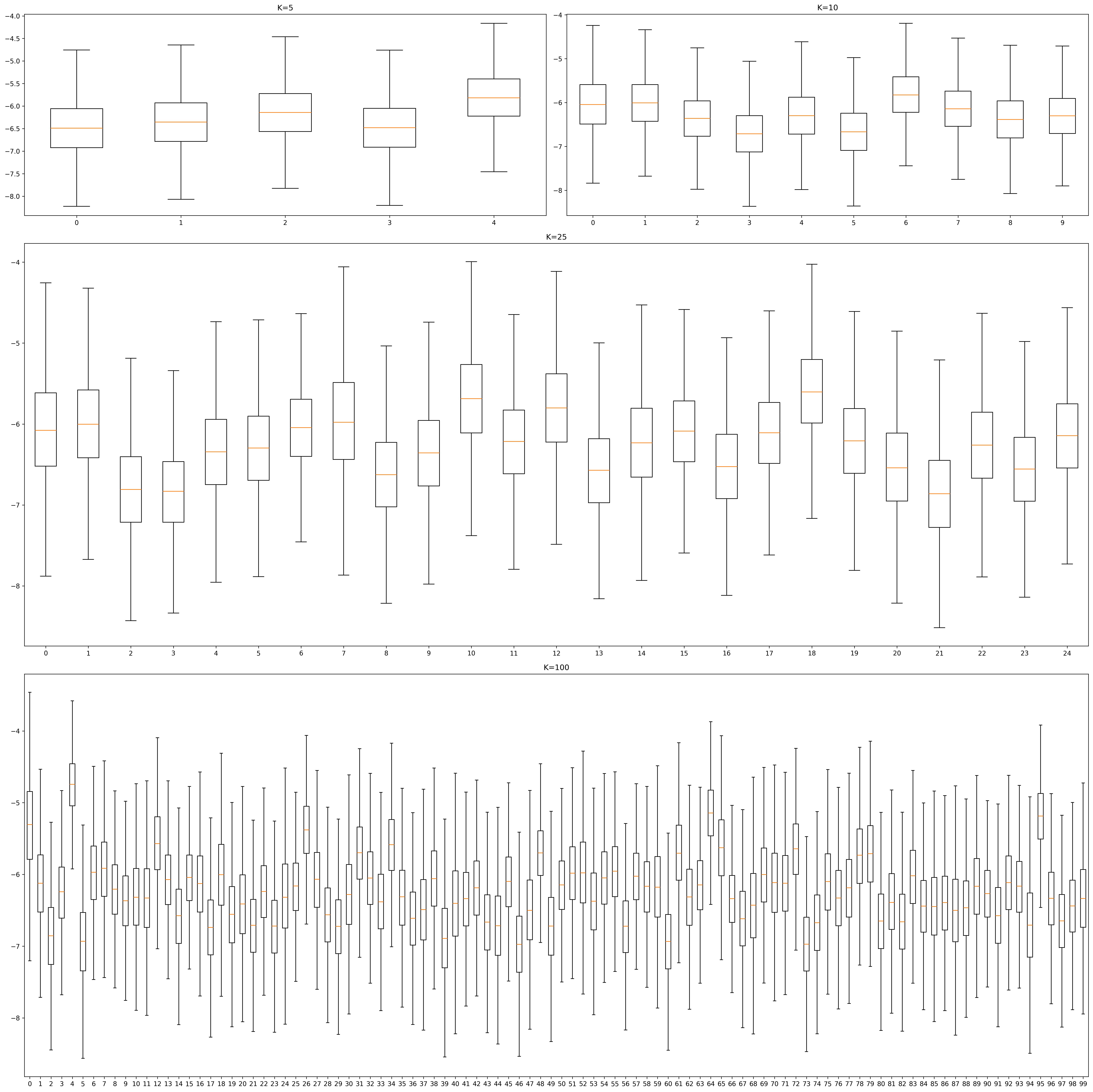}
    \caption{Distribution of docking scores for CKB-5M inside each cluster for different values of $K$. Outliers not shown.}
    \label{fig:scores_ckb-5m}
\end{figure}

\begin{figure}[H]
    \centering
    \includegraphics[width=1\linewidth]{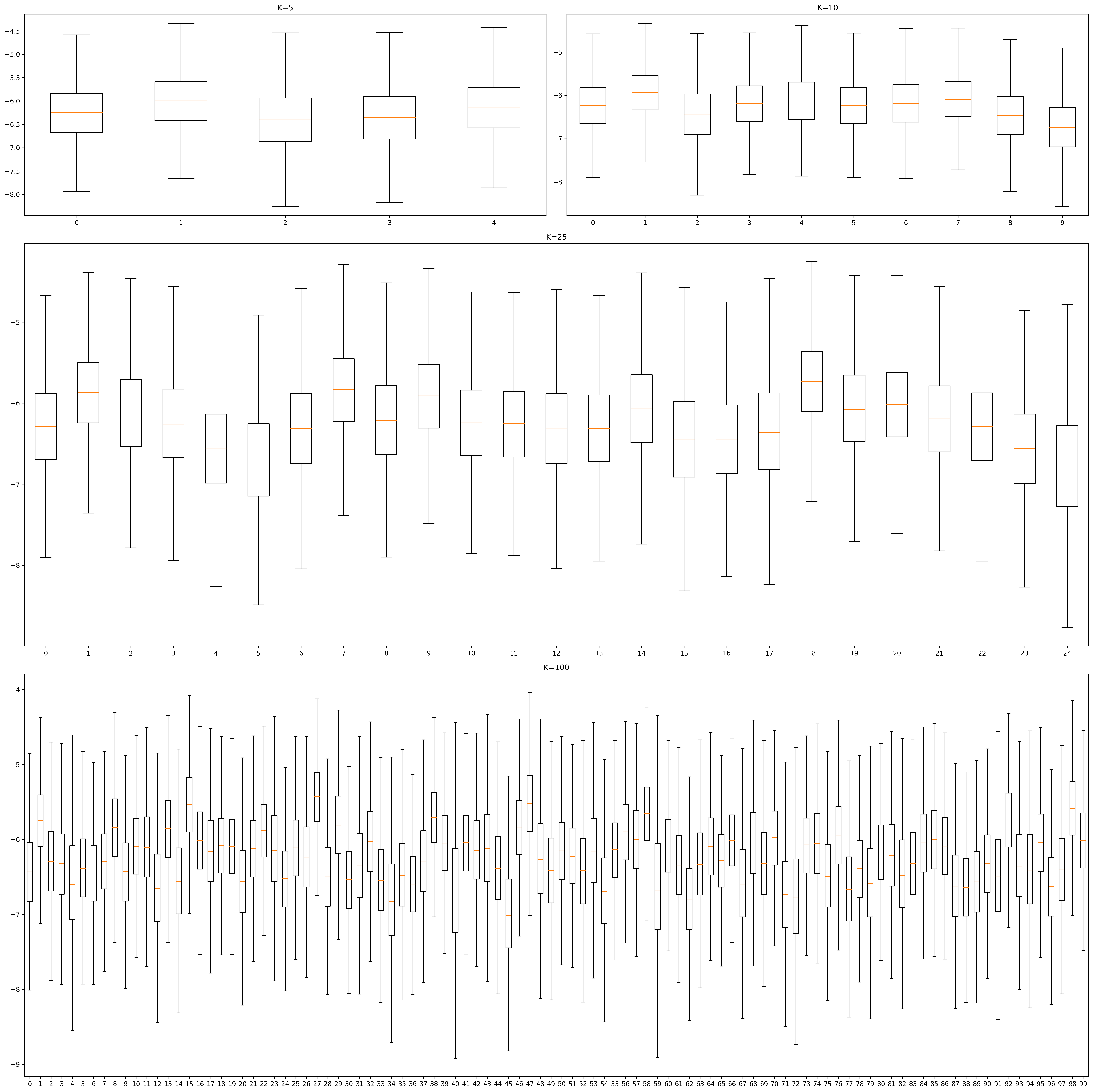}
    \caption{Distribution of docking scores for CKB-3.9M inside each cluster for different values of $K$. Outliers not shown.}
    \label{fig:scores_ckb-3.9m}
\end{figure}

\begin{figure}[H]
    \centering
    \includegraphics[width=1\linewidth]{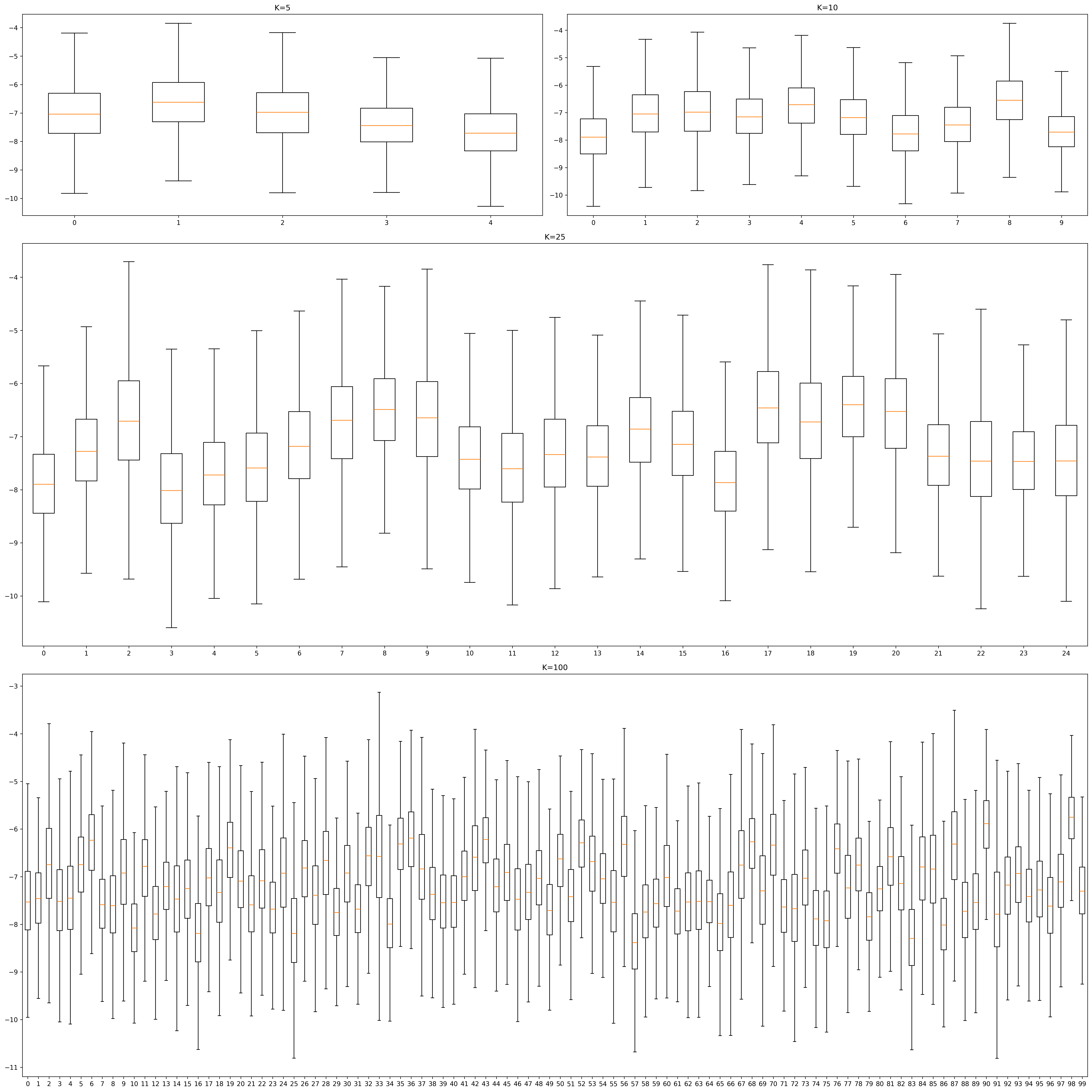}
    \caption{Distribution of docking scores for NEDD4-3.9M inside each cluster for different values of $K$. Outliers not shown.}
    \label{fig:scores_nedd4-3.9m}
\end{figure}

\newpage

\section{Physics-Based Docking} \label{docking-details}

We randomly selected 5 million compounds from an enumerated library of 69 billion compounds from \textsc{Enamine} \citep{Gorgulla2023}. 
In addition, an independent set of 3.9 million compounds was randomly sampled from \textsc{Enamine}’s S-class small-molecule database, which comprises approximately 3.9 billion unenumerated compounds, and retrieved in SMILES format. Only
one stereoisomer or one tautomer was retained for each SMILES string to preserve database diversity.
The compounds were then converted to three-dimensional structures and energy-minimized using RDKit \citep{rdkit} and Open Babel \citep{OBoyle2011}. 
The resulting library was subsequently docked against the target proteins with known active sites using the Uni-Dock small-molecule docking tool \citep{Yu2023} on NVIDIA RTX 4500 Ada Generation GPUs.

Specifically, we docked the libraries against the proteins CKB and NEDD4. 
CKB is a cytosolic phosphotransferase that buffers cellular energy by catalyzing the reversible transfer of a phosphate group between ATP and creatine, thereby helping maintain ATP levels during fluctuating energy demand. \citep{Bong2008}
NEDD4 is a HECT-family E3 ubiquitin ligase that recognizes substrates via WW domains and catalyzes ubiquitin transfer to regulate protein stability, trafficking, and signaling. \citep{Maspero2025}

We obtained docking scores against a Thymidylate Kinase (TMK) for the virtual library of 2M molecules from \textsc{Enamine}'s HTS database (\textsc{Enamine-HTS}) from \citep{Graff2021}. TMK is a phosphotransferase that catalyzes the reversible transfer of a phosphate group from ATP to deoxythymidine monophosphate (dTMP), to form deoxythymidine diphosphate (dTDP) in the process of DNA synthesis. \citep{Naik2015}

\end{document}